\documentclass[ss]{imsart}

\RequirePackage{amsthm,amsmath,amsfonts,amssymb}
\RequirePackage[numbers]{natbib}
\RequirePackage[colorlinks,citecolor=blue,urlcolor=blue]{hyperref}
\RequirePackage{graphicx}
\usepackage{multirow}
\usepackage{tabularx}
\usepackage{booktabs}
\usepackage{rotating}

\arxiv{2010.00000}
\startlocaldefs
\theoremstyle{plain}

\theoremstyle{definition}

\theoremstyle{remark}

\newcommand{\bftheta}{{\boldsymbol{\theta}}}

\newcommand{\bfx}{{\mathbf{x}}}

\newcommand{\xobs}{\boldsymbol{x}_{\mathrm{o}}}

\newcommand{\dd}{\mathrm{d}}

\endlocaldefs

\graphicspath{{Figures/}}

\begin{document}
\begin{frontmatter}
\title{A Review of Diffusion-based Simulation-Based Inference: Foundations and Applications in Non-Ideal Data Scenarios}
%\title{A sample article title with some additional note\thanksref{t1}}
\runtitle{Diffusion-based SBI: Foundations and Applications}
%\thankstext{T1}{A sample additional note to the title.}

\begin{aug}
%%%%%%%%%%%%%%%%%%%%%%%%%%%%%%%%%%%%%%%%%%%%%%%
%% Only one address is permitted per author. %%
%% Only division, organization and e-mail is %%
%% included in the address.                  %%
%% Additional information can be included in %%
\author[A]{\fnms{Haley}~\snm{Rosso}\ead[label=e1]{haley.rosso@emory.edu}\orcid{0009-0005-8583-4876}},
\and
\author[A]{\fnms{Talea}~\snm{Mayo}\ead[label=e1]{tlmayo@emory.edu}\orcid{0000-0002-7921-871X}}

%%%%%%%%%%%%%%%%%%%%%%%%%%%%%%%%%%%%%%%%%%%%%%
%% Addresses                                %%
%%%%%%%%%%%%%%%%%%%%%%%%%%%%%%%%%%%%%%%%%%%%%%
\address[A]{Department of Mathematics, Emory University, Atlanta, GA 30327, USA\printead[presep={,\ }]{e1}}

\end{aug}

\begin{abstract}
For complex simulation problems, inferring parameters often precludes the use of classical likelihood-based techniques due to intractable likelihoods. 
Simulation-based inference (SBI) methods offer a likelihood-free approach to directly learn posterior distributions $p(\bftheta \mid \xobs)$ from simulator outputs.
Recently, diffusion models have emerged as promising tools for SBI, addressing limitations of earlier neural methods such as neural likelihood/posterior estimation and normalizing flows.
This review examines diffusion-based SBI from first principles to applications, emphasizing robustness in three non-ideal data scenarios common to scientific computing: model misspecification (simulator-reality 
mismatch), unstructured or infinite-dimensional observations, and missing data.
We synthesize mathematical foundations and survey eight methods addressing these challenges, such as conditional diffusion for irregular data, guided diffusion for prior adaptation, sequential and factorized approaches for efficiency, and consistency models for fast sampling. 
Throughout, we maintain consistent notation and emphasize conditions required for accurate posteriors. We conclude with open problems and applications to geophysical uncertainty quantification, where these challenges are acute.
\end{abstract}


\end{frontmatter}

\newpage

\section{Introduction}\label{sec:intro}
Bayesian inference for complex scientific models increasingly relies on simulators that generate data through intricate numerical or stochastic pipelines but do not admit tractable likelihood functions.
This setting arises routinely in spatial statistics, epidemiology, geophysical modeling, and cosmology~\cite{hull_using_2022, gloeckler_all--one_2024, radev_bayesflow_2022, Wang__2023}.
While such simulators can produce realistic synthetic data, evaluating or even approximating likelihoods typically requires integrating over latent variables or invoking prohibitively expensive solvers.
As a result, classical likelihood-based Bayesian inference becomes infeasible.

Simulation-based inference (SBI) addresses this challenge by learning posterior distributions directly from simulator behavior rather than explicit likelihood evaluations.
In SBI, parameters are sampled from a prior, propagated through the simulator, and used to train a statistical or neural model capable of approximating the posterior distribution given new observations.
This paradigm has reshaped inference workflows across a wide range of scientific domains~\cite{cranmer_frontier_2020, lueckmann_benchmarking_2021}, enabling Bayesian analysis in scenarios previously considered computationally inaccessible.

Recent advances in score-based diffusion models provide a mathematically grounded and flexible framework for SBI.\
Rooted in score matching and stochastic calculus, diffusion models define generative processes via a forward noising mechanism and a reverse-time stochastic dynamics guided by learned score functions~\cite{ho_denoising_2020, song_generative_2020}.
These formulations admit both discrete and continuous-time perspectives and connect naturally to reverse-time diffusion, Fisher-divergence minimization, and stochastic differential equations~\cite{anderson_reverse-time_1982, hyvarinen_estimation_2005}.
In the context of SBI, diffusion models enable direct modeling of conditional posteriors without requiring invertible architectures or tractable Jacobians.

This flexibility contrasts with normalizing-flow-based approaches, which have been widely adopted for neural posterior and likelihood estimation.
Although normalizing flows allow exact likelihood evaluation via change-of-variables formulas, they impose architectural constraints through invertibility and Jacobian tractability, leading to trade-offs between expressiveness, stability, and computational cost~\cite{PracticalGuide}.
Diffusion models relax these constraints by relying only on score estimates and have demonstrated strong empirical performance in high-dimensional and complex inverse problems~\cite{song_generative_2020}, albeit often at increased sampling cost.

A central motivation for diffusion-based SBI lies in its potential to address non-ideal data regimes common in scientific applications.
First, \emph{model misspecification}---mismatch between the simulator and real-world data-generating processes---can lead to miscalibrated or overconfident posteriors, particularly for amortized neural inference methods~\cite{hermans_trust_2022, schmitt2024detectingmodelmisspecificationamortized, ward_robust_2022}.
Second, many observational datasets deviate from standard assumptions of fixed-dimensional, fully observed inputs, instead exhibiting missing values, irregular sampling, or function-valued structure.
Such settings arise in climate monitoring, ecological time series, and PDE-driven inverse problems, where parameters or observations naturally live in high- or infinite-dimensional spaces~\cite{gloeckler_all--one_2024, baldassari_conditional_2023}.
These challenges strain traditional amortization strategies and motivate inference methods capable of accommodating unstructured data and discretization-dependent representations.

A growing body of work bridges diffusion models and SBI, developing conditional and sequential score-based approaches for posterior inference.
Notable examples include neural posterior score estimation and its sequential variants~\cite{geffner2023compositionalscoremodelingsimulationbased, sharrock2024sequentialneuralscoreestimation}, as well as architectures inspired by conditional diffusion models for inverse problems and data imputation~\cite{tashiro_csdi_2021, gloeckler_all--one_2024}.
Parallel research investigates diffusion models in function spaces and discretization-invariant formulations, highlighting foundational questions at the interface of stochastic analysis, numerical discretization, and scientific computing~\cite{baldassari_conditional_2023, pidstrigach2025infinitedimensionaldiffusionmodels}.

Despite rapid progress, diffusion-based SBI remains an evolving field with several open challenges.
These include robustness under simulator misspecification, principled treatment of infinite-dimensional parameters, computational efficiency in sequential settings, and mechanisms for incorporating evolving or data-driven priors.
This review synthesizes the foundations and recent advances in diffusion-based SBI, with a particular emphasis on these challenges, and highlights directions for future research relevant to uncertainty quantification and computational statistics.

\subsection{Contributions}
This article is a survey/review and introduces no new algorithms.
The review focuses exclusively on diffusion models for simulation-based inference, with two primary goals:
\begin{enumerate}
    \item to present a rigorous synthesis of the theoretical foundations underpinning diffusion-based SBI, including score matching, reverse-time dynamics, discretization effects, and calibration; and
    \item to survey existing algorithms and applications while articulating open problems related to misspecification, identifiability, function-space posteriors, and computational efficiency.
\end{enumerate}
Closely related approaches, such as flow matching, are discussed only when useful for contrast.

\subsection{Outline}
Section~\ref{sect:SBI} introduces simulation-based inference and its foundations in Bayesian statistics, establishing notation and reviewing classical and neural SBI paradigms.
Section~\ref{sect:diffusion_models} develops the diffusion-model background required for SBI, including forward processes, reverse-time dynamics, and score matching.
Section~\ref{sect:data_limitations} examines non-ideal data regimes—misspecification, missing data, and unstructured or infinite-dimensional observations—and their implications for amortized inference.
Section~\ref{sect:architectures} reviews diffusion-based SBI architectures designed to address these challenges.
Section~\ref{sect:lit_survey} synthesizes recent literature with an emphasis on methodological design choices.
The paper concludes in section~\ref{sect:conclusion} with open problems and future directions motivated by applications in uncertainty quantification.

\section{Chronology of SBI}\label{sect:SBI}
We briefly review the development of simulation-based inference (SBI) to contextualize diffusion-based approaches.
Comprehensive treatments of SBI are available elsewhere~\cite{cranmer_frontier_2020, lueckmann_benchmarking_2021, falkiewicz2023calibratingneuralsimulationbasedinference, sharrock2024sequentialneuralscoreestimation, kelly_simulation-based_2025}; our focus here is to establish notation and highlight conceptual milestones relevant to diffusion-based methods.

\subsection{Problem setup and notation}\label{prob_and_notation}

Let $\Theta$ denote the parameter space and $\mathcal{X}$ the observation space.
We write $\bftheta \in \Theta$ for unknown parameters and $\bfx_{\text{o}} \in \mathcal{X}$ for observed data, while $\bfx$ denotes simulated data generated by a simulator.
Throughout, $p(\cdot)$ denotes probability densities or measures, with $p(\bftheta)$ the prior, $p(\bfx \mid \bftheta)$ the likelihood, and $p(\bftheta \mid \bfx_{\text{o}})$ the posterior.

In many scientific applications, $\Theta$ and $\mathcal{X}$ need not be finite-dimensional vector spaces.
Parameters may be function-valued or time-dependent, and observations may be irregularly sampled, partially observed, or defined through measurement operators.
These features complicate inference and motivate SBI methods capable of accommodating unstructured or infinite-dimensional data representations.
The nuances of the parameter space depend heavily on the specific scientific application and model structure; through this review, we will return to this point in more detail (cf.~sections~\ref{subsect:unstruct_data},~\ref{sect:baldassari}).

\subsection{Bayesian inference}\label{sect:bayes_inference}
Bayesian inference aims to compute the posterior distribution $p(\bftheta \mid \bfx_{\text{o}})$ given parameters $\bftheta$ and observed data $\bfx_{\text{o}}$. 
By Bayes' theorem,
\begin{equation}\label{eq:bayes_thm}
    p(\bftheta \mid \bfx_{\text{o}}) =
    \frac{p(\bfx_{\text{o}} \mid \bftheta)\,p(\bftheta)}
    {\int_\Theta p(\bfx_{\text{o}} \mid \bftheta')\,p(\bftheta')\,\dd\bftheta'},
\end{equation}
where the denominator is the marginal likelihood (evidence), which is generally intractable in simulation-based settings~\cite{falkiewicz2023calibratingneuralsimulationbasedinference,sharrock2024sequentialneuralscoreestimation,Wang__2023}.

In SBI, the likelihood $p(\bfx \mid \bftheta)$ cannot be evaluated and is accessible only through a simulator $g(\bftheta,\bfx)$ that generates samples $\bfx \sim p(\bfx \mid \bftheta)$ from prior draws $\bftheta \sim p(\bftheta)$. 
This black-box structure precludes classical MCMC and variational inference~\cite{cranmer_frontier_2020,boelts_flexible_2022} and motivates likelihood-free approaches.

Priors in SBI are typically informed by scientific domain knowledge rather than conjugacy, yet remain a major source of model misspecification. 
Overly broad priors reduce efficiency, while overly restrictive priors risk excluding plausible parameter values~\cite{kelly_simulation-based_2025,boelts_flexible_2022}.

Beyond these components, SBI methods share several defining features:
\begin{enumerate}
    \item the simulator is treated as a \textbf{black box} mapping parameters to data;
    \item inference is \textbf{amortized}, with models trained offline to enable fast inference on new observations;
    \item high-dimensional data are compressed via \textbf{summary statistics} designed to retain posterior-relevant information.
\end{enumerate}

\subsection{Classical SBI:\ ABC and synthetic likelihood}\label{sect:classical_SBI}

One of the earliest and most widely used simulation-based inference methods is Approximate Bayesian Computation (ABC)~\cite{lueckmann_benchmarking_2021,ward_robust_2022,kelly_simulation-based_2025}. 
ABC samples parameters \(\bftheta \sim p(\bftheta)\), simulates data \(\bfx \sim p(\bfx \mid \bftheta)\), and accepts \(\bftheta\) if a discrepancy \(\rho(S(\bfx), S(\xobs)) < \varepsilon\) holds for chosen summary statistics \(S(\cdot)\) and tolerance \(\varepsilon\)~\cite{papamakarios_fast_2018}. 
While conceptually simple and broadly applicable, ABC scales poorly: acceptance rates decay rapidly with dimension, making inference prohibitively expensive as \(\varepsilon \to 0\).

Synthetic likelihood (SL) methods~\cite{price2018bayesian,papamakarios_fast_2018} address this limitation by approximating the intractable likelihood \(p(\xobs \mid \bftheta)\) directly.
At each \(\bftheta\), a parametric density (typically Gaussian) is fitted to summary statistics computed from repeated simulations, replacing the hard acceptance step of ABC with a smooth likelihood approximation. 
This enables standard MCMC but retains dependence on informative summaries and struggles in high-dimensional or multimodal settings.

More recent synthetic neural likelihood (SNL) approaches replace parametric likelihood models with neural networks, increasing expressivity while preserving the simulation-based structure. 
As such, SL and SNL form a bridge between classical ABC and modern neural likelihood estimation methods, which we review next.

\subsection{Neural SBI methods}
Neural simulation-based inference (SBI) uses neural density estimators trained on simulated pairs \(\{(\bftheta_i,\bfx_i)\}_{i=1}^N \sim p(\bftheta)p(\bfx\mid\bftheta)\) to enable flexible, amortized inference in high-dimensional settings. 
Compared to classical SBI, neural methods can scale better to high-dimensional data, learn complex posterior geometries, and leverage modern deep learning techniques for improved performance.
Neural SBI methods are commonly categorized by the posterior-relevant quantity they approximate.

Neural likelihood estimation (NLE) learns a surrogate likelihood model and enables posterior inference via standard sampling methods.
Neural ratio estimation (NRE) estimates the likelihood-to-evidence ratio using classification-based objectives, yielding posterior densities up to normalization~\cite{hermans_trust_2022}.
Neural posterior estimation (NPE) directly approximates the posterior distribution, enabling fully amortized inference without per-observation sampling~\cite{papamakarios_fast_2018, hermans_trust_2022,lueckmann_benchmarking_2021}.

Each of these three methods have their own benefits and drawbacks, and can be combined in various ways to leverage their individual strengths~\cite{geffner2023compositionalscoremodelingsimulationbased,simons_simulation-based_nodate}.
NPE offers fast inference at the cost of sensitivity to proposal mismatch, while NLE and NRE provide robustness to changing priors but require sampling at inference time.
Sequential variants of all three approaches improve simulation efficiency by adaptively refining proposals, at the expense of full amortization~\cite{papamakarios_sequential_2019}.
The choice among these methods depends on the inference task, computational budget, and degree of reuse across observations.

\subsection{Limitations of flow-based NPE}\label{sect:flow_based_limits}

Normalizing-flow-based models are widely used in neural posterior estimation due to their expressive power and exact density evaluation~\cite{dax_flow_2023}.
However, flow architectures impose structural constraints through invertibility and tractable Jacobians, limiting expressiveness and complicating training for sharp, multimodal, or heavy-tailed posteriors.
Moreover, flow-based models typically require fixed-dimensional inputs, making them ill-suited to irregular observations, missing data, or function-valued parameters without lossy preprocessing~\cite{chen_conditional_2025}.

These limitations motivate diffusion-based SBI methods, which replace density evaluation with score estimation, relax architectural constraints, and offer greater flexibility for conditioning on unstructured or high-dimensional data~\cite{sharrock2024sequentialneuralscoreestimation,geffner2023compositionalscoremodelingsimulationbased}.
The connection between diffusion models and SBI is developed in the following section.

\section{Diffusion model background}~\label{sect:diffusion_models}
Diffusion models are generative models that learn complex target distributions through a noising-denoising construction.
A \emph{forward} process gradually perturbs data into a tractable reference distribution (typically Gaussian), while a \emph{reverse-time} process generates samples by approximately inverting this perturbation using a learned \emph{score function}.
Modern score-based diffusion models unify discrete diffusion probabilistic models and continuous-time stochastic formulations; see, e.g.,~\cite{sohldickstein2015deepunsupervisedlearningusing, ho_denoising_2020, song_generative_2020, song_score-based_2021}.

While diffusion models are a relatively recent development in machine learning, the foundational concepts date back several decades.
The reverse generative process dates back to the early 1980s with~\cite{anderson_reverse-time_1982}'s work on the time reversal of diffusion processes, which shows that the time reversal of diffusion processes is itself a diffusion process, with the drift term depending on the score function (the gradient of the log-density of the noised data)~\cite{song_score-based_2021}.
Fundamental techniques like denoising score matching, introduced by~\cite{hyvarinen_estimation_2005} and elaborated on by~\cite{vincent2011connection}, enable the estimation of this score function using a neural network without direct density calculation.

Later,~\cite{sohldickstein2015deepunsupervisedlearningusing} proposed a discrete-time diffusion process that gradually transforms data into noise through a series of small Gaussian perturbations.
This 2015 is often cited as one of the earliest introductions of diffusion probabilistic models.
The authors framed diffusion models from a thermodynamic perspective, using non-equilibrium thermodynamics to define a forward noising process and a corresponding reverse generative process.
These results ultimately led to the development of score-based generative models, which use score matching to learn the gradients of the data distribution and constitute the main focus of this review~\cite{song_generative_2020, song_score-based_2021}.

Subsequently, modern diffusion models achieved prominence through two parallel developments that combined earlier diffusion process ideas with modern deep learning architectures.
In 2019,~\cite{song_generative_2020} introduced score-based generative modeling by estimating gradients of the data distribution, which uses Langevin dynamics to sample from a sequence of decreasing noise scales.
Shortly thereafter,~\cite{ho_denoising_2020} introduced denoising diffusion probabilistic models (DDPM) in 2020. 
DDPMs train a sequence of probabilistic models to reverse each step of the noise corruption and demonstrated remarkable success in generating high-quality samples.

The work of~\cite{ho_denoising_2020} significantly popularized diffusion models by providing a simpler training objective and demonstrating impressive generative capabilities, especially in image generation, and is a cornerstone for elucidating modern diffusion models.
Notably, both~\cite{song_generative_2020} and~\cite{song_score-based_2021} established the connection between diffusion processes and score-based generative models, effectively unifying multiple existing approaches and demonstrating the usefulness of score matching in generative modeling.

Initially, the two main branches of these models, Denoising Diffusion Probabilistic Models (DDPM) and Score Matching with Langevin Dynamics (SMLD), were treated as discrete-time frameworks. 
Song~\cite{song_score-based_2021} showed that these discrete models are actually discretizations of underlying SDEs; namely, they identified that SMLD converges to a variance exploding (VE) SDE, while DDPM converges to a variance preserving (VP) SDE.\

The field coalesced and accelerated significantly with the introduction of continuous-time formulations.
While the foundational breakthrough for this class of models occurred in a discrete-time setting,~\cite{song_score-based_2021} generalized the concept of noise scales by treating them as a continuum of distributions evolving over time according to a prescribed SDE.\
The fundamental mathematical result that allows for continuous-time diffusion models was established by~\cite{anderson_reverse-time_1982}, who showed that the time reversal of diffusion processes is itself a diffusion process, with the drift term depending on the score function (the gradient of the log-density of the noised data)~\cite{song_score-based_2021}.

This new idea brought forth the probability flow ODE (cf.~section~\ref{sect:probflow}), which shares the same marginal distributions as the SDE process. 
The probability flow ODE enabled deterministic sampling and faster adaptive sampling.

The SDE framework introduced by Song et al.~also showed that the \emph{conditional} reverse-time SDE could be estimated from \emph{unconditional} scores.
This result paved the way for powerful guidance mechanisms to emerge, such as classifier guidance~\cite{dharwalnichol} and classifier-free guidance~\cite{ho2022classifierfreediffusionguidance}, allowing diffusion models to solve inverse problems or perform conditional generation without specific retraining (cf.~section~\ref{sect:guided_diffusion}).
Further practical improvements were achieved by~\cite{karras2022elucidatingdesignspacediffusionbased}, who clarified the design space, leading to improvements in sampling processes (e.g., higher-order ODE solvers) and achieving state-of-the-art results with significantly fewer network evaluations (e.g., 35 evaluations per image).

These advancements solidified diffusion models as a powerful and flexible class of generative models, notably for their superior sample quality compared to earlier methods like generative adversarial networks (GANs) and their ability to use unconstrained architectures, unlike normalizing flows.
With these works in mind, we will review the mathematical foundations behind diffusion models in the following sections, first establishing notation before proceeding to the technical details.

\smallskip

\noindent \textbf{Notation.} Let $\bfx_0 \sim p_{\text{data}}$ denote a data sample and $\{\bfx_t\}_{t\in[0,T]}$ its noised versions under the forward process.
We write $p_t(\bfx_t)$ for the forward marginal at time $t$ and $\nabla_{\bfx_t}\log p_t(\bfx_t)$ for its score.
In SBI contexts, the diffusion ``state'' $\bfx$ may represent parameters, observations, or summary representations, but we retain $\bfx_t$ to match diffusion conventions.

\subsection{Forward process}\label{sect:fwd_process}
The forward process corrupts structured data into noise, typically a standard Gaussian.
Both discrete-time and continuous-time formulations share this goal: starting from $\bfx_0 \sim p_{\text{data}}$, the process destroys structure over time $t \in [0,T]$ until $\bfx_T \approx \mathcal{N}(0,I)$.

In continuous time, it is modeled by an It\^o SDE~\cite{song_score-based_2021}:
\begin{equation}\label{eq:itoSDE}
    d\bfx_t = f(\bfx_t,t)\,dt + g(t)\,d\mathbf{W}_t, 
    \qquad \bfx_0 \sim p_{\text{data}}.
\end{equation}
A widely used choice is the variance-preserving (VP) SDE (corresponding to DDPM in the small-step limit):
\begin{equation}\label{eq:var_preserving}
    d\bfx_t = -\tfrac{1}{2}\beta(t)\bfx_t\,dt + \sqrt{\beta(t)}\,d\mathbf{W}_t,
\end{equation}
which drives $\bfx_t$ toward a Gaussian reference distribution as $t\to T$.
Discrete-time diffusion models (e.g., DDPM) can be viewed as numerical discretizations (e.g., Euler-Maruyama) of such SDEs~\cite{song_score-based_2021, ho_denoising_2020,sohldickstein2015deepunsupervisedlearningusing} and remain popular for their simplicity and stable training~\cite{dharwalnichol,batzolis_conditional_2021,rasul2021autoregressivedenoisingdiffusionmodels}.
Explicit formulations of the discrete-time forward process can be found in Appendix~\ref{app:discrete_diffusion_fwd}.

While discrete models suffice for many applications, the continuous SDE framework~\cite{song_score-based_2021} brought forth three key advances:
\begin{enumerate}
    \item \textbf{Probability flow ODEs} (cf.~section~\ref{sect:probflow}): A deterministic counterpart to~\eqref{eq:itoSDE} with identical marginals, enabling exact likelihood evaluation and efficient adaptive solvers that reduce network evaluations by 90\%~\cite{song_score-based_2021}.
    \item \textbf{Flexible discretization}: Any numerical SDE/ODE solver applies; practitioners can trade accuracy for speed at inference time without retraining.
    \item \textbf{Predictor-corrector samplers} (cf.~section~\ref{sect:annealed}): Hybrid schemes alternating deterministic ODE steps with Langevin refinement improve sample quality by 10--30\%~\cite{song_score-based_2021, gloeckler_all--one_2024}.
\end{enumerate}
Key foundational papers from~\cite{sohldickstein2015deepunsupervisedlearningusing},~\cite{ho_denoising_2020},~\cite{song_generative_2020}, and~\cite{song_score-based_2021} established these concepts and should be referred to for further detail. 
The principles of discrete and continuous-time settings also broadly applies to the reverse process; a more rigorous discussion of this will be provided in the following section.

\subsection{Reverse-time SDE}\label{sect:reversetime_SDE}

Sampling is performed by simulating a reverse-time process from $\bfx_T\sim p_T(\bfx_T)\approx\mathcal{N}(0,I)$ back to $t=0$.
A key mathematical result is that time reversal of a diffusion remains a diffusion, with a drift correction involving the score~\cite{anderson_reverse-time_1982, song_score-based_2021}.
Specifically, the reverse-time SDE associated with~\eqref{eq:itoSDE} is
\begin{equation}\label{eq:reverse_sde}
    d\bfx_t
    =
    \big[
        f(\bfx_t,t) - g(t)^2 \nabla_{\bfx_t}\log p_t(\bfx_t)
    \big]\,dt
    + g(t)\,d\bar{\mathbf{W}}_t,
\end{equation}
where $\bar{\mathbf{W}}_t$ is a Wiener process in reverse time.
Because $p_t(\bfx_t)$ is intractable, one trains a neural score model $s_\phi$ so that
$s_\phi(\bfx_t,t)\approx \nabla_{\bfx_t}\log p_t(\bfx_t)$ and substitutes it into~\eqref{eq:reverse_sde} to enable practical sampling.

\subsubsection{Probability flow ODE}\label{sect:probflow}
In addition to the stochastic reverse SDE, there exists a deterministic ODE with the same time-marginals~\cite{song_score-based_2021}:
\begin{equation}\label{eq:probflow_ODE}
     \frac{d\bfx_t}{dt}
     =
     f(\bfx_t, t)
     - \tfrac{1}{2} g(t)^2 \nabla_{\bfx_t}\log p_t(\bfx_t).
\end{equation}
The probability flow ODE enables deterministic sampling and supports likelihood evaluation via instantaneous change-of-variables, while in practice many samplers trade accuracy for speed using numerical solvers and discretization choices~\cite{song_score-based_2021, karras2022elucidatingdesignspacediffusionbased}.

\subsection{Sampling: annealed Langevin dynamics}\label{sect:annealed}
A common discrete-time score-based sampler is \emph{annealed Langevin dynamics} (ALD), which updates samples using the score at a sequence of decreasing noise levels~\cite{song_generative_2020}.
ALD builds upon standard Langevin dynamics (LD), which iteratively refines an initial random sample by moving it toward regions of higher probability density.

In a generic form,
\begin{equation}\label{eq:ald_update}
    \bfx \leftarrow \bfx + \delta_t\, s_\phi(\bfx,t,c) + \sqrt{2\delta_t}\,\boldsymbol{\eta}_t,
    \qquad \boldsymbol{\eta}_t\sim\mathcal{N}(0,I),
\end{equation}
where $\delta_t$ is a step size and $c$ denotes optional conditioning.
In continuous-time implementations, ALD often appears as a \emph{corrector} within predictor-corrector samplers, refining proposals produced by an SDE/ODE predictor step~\cite{song_score-based_2021}.

The probability flow ODE and ALD sampler concepts round out the reverse-time process in continuous diffusion models.
However, a critical component remains: training the score network to accurately estimate the score function across noise levels.
This is the focus of the next section.

\subsection{Score matching}\label{subsubsect:score_match}
Score matching is the training principle underlying score-based generative models, enabling approximation of the score function used in the reverse diffusion process~\cite{song_pseudoinverse-guided_2022,rozet_score-based_2023,sharrock2024sequentialneuralscoreestimation}.
The score is defined as the gradient of the log-density,
\begin{equation}\label{eq:score_fcn}
    \nabla_{\bfx}\log p(\bfx),
\end{equation}
and is approximated by a neural score network
\begin{equation}\label{eq:score_net}
    s_\phi(\bfx_t,t),
\end{equation}
where $\bfx_t$ denotes data corrupted by the forward diffusion process at time $t$.
Training is typically performed via \emph{denoising score matching} (DSM), which minimizes
\begin{equation}\label{eq:DSM_Loss}
    \mathbb{E}_{\bfx_0,t,\bfx_t}\!
    \left[
    \sigma^2(t)\,
    \big\|
    s_\phi(\bfx_t,t)
    - \nabla_{\bfx_t}\log p_t(\bfx_t\mid\bfx_0)
    \big\|_2^2
    \right],
\end{equation}
where $p_t(\bfx_t\mid\bfx_0)=\mathcal{N}(\mu(t)\bfx_0,\sigma^2(t)I)$ denotes the forward noising kernel.

In practice, DSM is often reparameterized as a {noise prediction} objective for numerical stability~\cite{rozet_score-based_2023}. 
A common equivalent parameterization predicts the injected noise $\boldsymbol{\epsilon}$ for numerical stability~\cite{rozet_score-based_2023}:
\begin{equation}\label{eq:DSM_loss_reparam}
    \mathbb{E}_{\bfx_0,t,\boldsymbol{\epsilon}}
    \big[
    \|\boldsymbol{\epsilon}_\phi(\bfx_t,t)-\boldsymbol{\epsilon}\|_2^2
    \big],
\end{equation}
using $\bfx_t=\mu(t)\bfx_0+\sigma(t)\boldsymbol{\epsilon}$ with $\boldsymbol{\epsilon}\sim\mathcal{N}(0,I)$.
Once trained, $s_\phi$ (or $\boldsymbol{\epsilon}_\phi$) can be inserted into the reverse dynamics~\eqref{eq:reverse_sde} or~\eqref{eq:probflow_ODE} to generate samples.

Together with the forward process, score matching enables learning a score function across all noise levels, forming the foundation for reverse-time sampling via SDEs or probability flow ODEs.
This mechanism is central to diffusion-based generative modeling and underpins its use in simulation-based inference.

\subsection{Connection to SBI}\label{sect:diffusion_SBI_connection}
Score matching provides a direct bridge from diffusion models to likelihood-free inference~\cite{hyvarinen_estimation_2005, vincent2011connection}.
In diffusion-based SBI, one trains a conditional score model for the posterior,
\[
s_\psi(\bftheta,t,\bfx_{\mathrm{o}})
\approx
\nabla_\bftheta \log p_t(\bftheta\mid\bfx_{\mathrm{o}}),
\]
and then generates posterior samples by integrating a corresponding reverse-time SDE/ODE in parameter space.
This viewpoint underlies neural posterior score estimation methods that directly model conditional posterior scores and enable flexible conditioning at inference time~\cite{geffner2023compositionalscoremodelingsimulationbased, sharrock2024sequentialneuralscoreestimation}.

An alternative strategy models the joint distribution $p(\bftheta,\bfx)$ with a single score network, from which conditionals (posterior, likelihood, predictive) can be obtained via score-based conditioning; Simformer is a representative example designed for unstructured and missing observations~\cite{gloeckler_all--one_2024}.
To incorporate multiple observations, composition mechanisms exploit score additivity to aggregate evidence at inference time~\cite{geffner2023compositionalscoremodelingsimulationbased}.
Sequential diffusion-based SBI methods refine proposals across rounds to allocate simulations adaptively to posterior-relevant regions~\cite{sharrock2024sequentialneuralscoreestimation}.

\begin{figure}
    \centering
    \includegraphics[width=\linewidth]{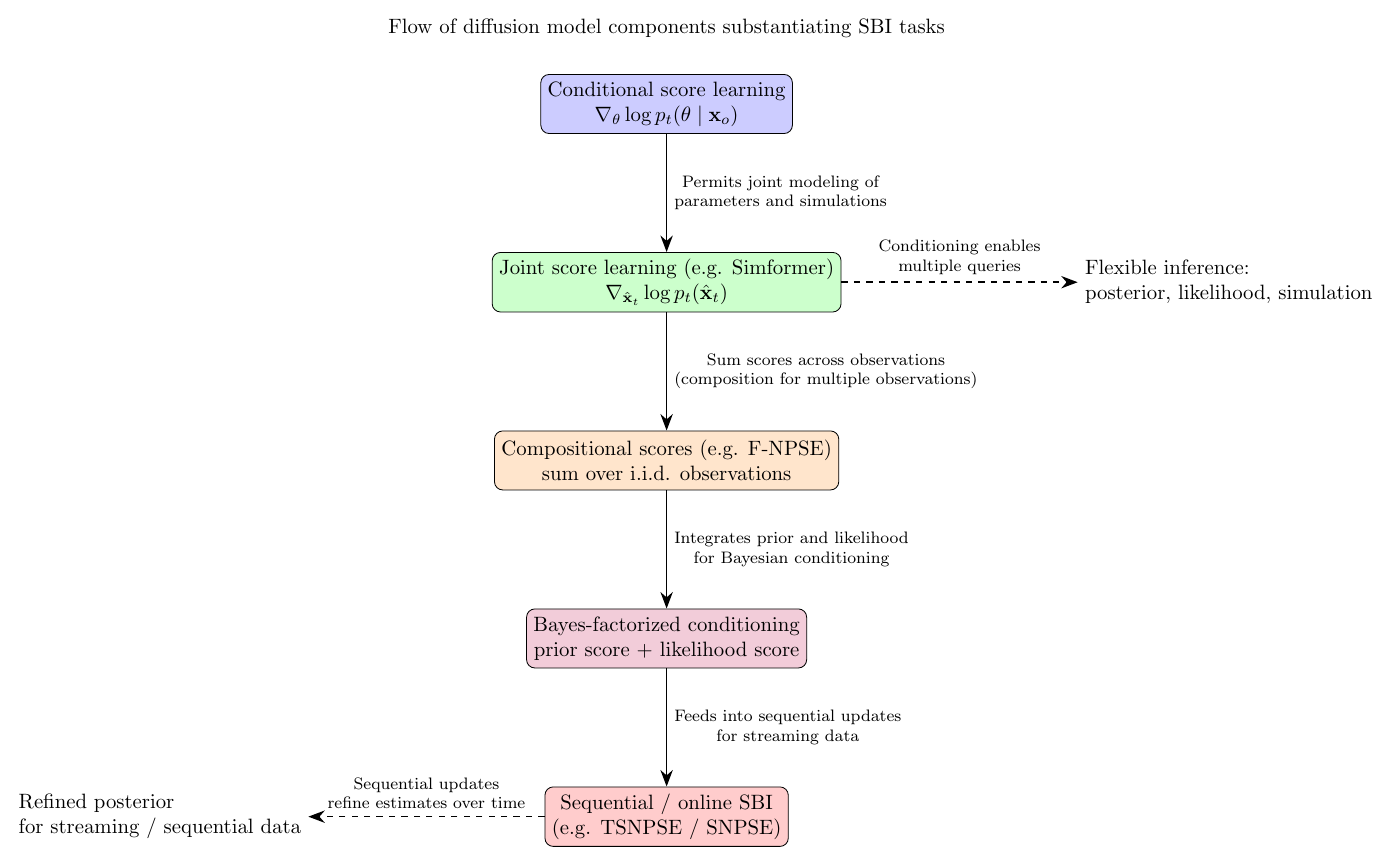}
    \caption{Conceptual connection between diffusion models and simulation-based inference (SBI). Score matching enables estimation of intractable score functions, which define reverse-time dynamics used for conditional generation and posterior sampling}\label{fig:diffusion-sbi-flowchart}
\end{figure}

These approaches demonstrate score matching's relevance for diffusion-based SBI.\
Figure~\ref{fig:diffusion-sbi-flowchart} summarizes how forward noising, score matching, and reverse-time dynamics connect diffusion models to SBI objectives.

\section{Limitations of SBI with data}\label{sect:data_limitations}
Real-world SBI applications often violate the idealized assumptions under which amortized neural inference is trained.
We focus on three interrelated data challenges that can compromise posterior reliability: \emph{missing observations}, \emph{unstructured/irregular data}, and \emph{model misspecification}.
Classical SBI (e.g., ABC and flow-based NPE) can struggle in these regimes, motivating diffusion-based architectures that offer greater flexibility for conditioning and representation, at the cost of additional computation.
This section formalizes the three challenges and sets evaluation criteria used in our literature survey (section~\ref{sect:lit_survey}).

\subsection{Missing data}\label{subsect:missing_data}
Let $\bfx\in\mathbb{R}^d$ be a data vector and let $s\in\{0,1\}^d$ be a binary mask with $s_i=1$ if $(\bfx)_i$ is observed and $s_i=0$ if missing.
Write $\bfx=(\bfx_{\mathrm{obs}},\bfx_{\mathrm{miss}})$, where only $\bfx_{\mathrm{obs}}$ is available.
Common missingness mechanisms include MCAR, MAR, and MNAR~\cite{mcknight_missing_2007, joel_review_2022}:
\begin{itemize}
    \item MCAR:\ $P(s\mid \bfx)=P(s)$,
    \item MAR:\ $P(s\mid \bfx)=P(s\mid \bfx_{\mathrm{obs}})$,
    \item MNAR:\ $P(s\mid \bfx)$ depends on $\bfx_{\mathrm{miss}}$.
\end{itemize}
For the scope of this review, we consider this to be distinct from unstructured data settings, and these differences are delineated in section~\ref{subsect:unstruct_data}.
See Appendix~\ref{appendix:missing_data} for formal definitions.

Most neural SBI pipelines assume fixed-dimensional inputs and are not trained to marginalize missing values~\cite{liu_physics-inspired_2024, radev_bayesflow_2022}.
Naive preprocessing (zero filling, mean imputation, or dropping entries) yields a learned conditional $q_\psi(\bftheta\mid \tilde{\bfx})$ that generally targets the wrong posterior, since the correct quantity is
\begin{equation}\label{eq:missing_marginalization}
    p(\bftheta\mid \bfx_{\mathrm{obs}})
    =
    \int p(\bftheta\mid \bfx_{\mathrm{obs}},\bfx_{\mathrm{miss}})
    \,p(\bfx_{\mathrm{miss}}\mid \bfx_{\mathrm{obs}})\,d\bfx_{\mathrm{miss}}.
\end{equation}
Without modeling (or integrating out) $\bfx_{\mathrm{miss}}$, posterior uncertainty is often miscalibrated and estimates can be biased~\cite{verma_robust_2025, wang_missing_2024}.

\subsection{Model misspecification}\label{sect:model_misspec}
Misspecification refers to discrepancies between the simulator and the real data-generating process.
In SBI this can arise through (i) structural mismatch ($p(\bfx\mid\bftheta)$ cannot represent the true mechanism), (ii) prior-data conflict (the prior assigns low mass near the data-consistent parameter region), or (iii) observations that lie outside the distribution induced by the training simulations (often described as out-of-distribution or ``out-of-simulation'' behavior)~\cite{ward_robust_2022, kelly_misspecification-robust_2024, schmitt2024detectingmodelmisspecificationamortized}.

Neural SBI methods learn $q_\psi(\bftheta\mid\bfx)\approx p(\bftheta\mid\bfx)$ under the training joint distribution $p(\bftheta)p(\bfx\mid\bftheta)$.
When $\bfx_{\mathrm{o}}$ exhibits patterns not represented by this training distribution, amortized estimators extrapolate poorly and can produce overconfident or biased posteriors~\cite{kelly_simulation-based_2025, elsemuller_sensitivity-aware_2024, schmitt2024detectingmodelmisspecificationamortized}.
Flow-based NPE is particularly vulnerable because its density-matching
objective amplifies errors in low-probability tail regions.

\noindent\textbf{Intuition and real-world example.} There are many real-world scenarios that can describe the intuition behind model misspecification in SBI.\
Here, we present an example for surge modeling, which is described in our future work section (cf.~section~\ref{sect:conclusion}) as an application we are actively pursuing to resolve these challenges. 

A coastal flood simulator calibrated on historical storms may underrepresent tail dynamics (e.g., extreme winds or compound flooding).
For a rare Category~5 event, an amortized posterior can confidently infer parameters that reproduce the observed pattern poorly (e.g., systematically underestimating peak surge), yielding a ``silent failure'' unless diagnostics are performed.

Diffusion-based approaches can mitigate some aspects of misspecification (e.g., by enabling inference-time guidance or prior adaptation), but no method fully resolves cases where the simulator is structurally wrong; diagnostics and robustness checks remain essential.

\subsection{Unstructured data}\label{subsect:unstruct_data}

Unstructured data lack a common fixed ambient dimension: observations may have varying length, resolution, support, or measurement geometry~\cite{shukla_survey_2021, gloeckler_all--one_2024}.
Examples include irregular time series, sets of sensor readings at varying locations, fields represented on different meshes, and heterogeneous modalities (e.g., point clouds or function-valued parameters)~\cite{gloeckler_all--one_2024, baldassari_conditional_2023}.
Unlike missingness, unstructured data may not correspond to a masked subset of a shared $\mathbb{R}^d$; rather, each sample may live in a different space $\mathcal{X}_i$ and correspond to different measurement operators.

\begin{figure}[t]
    \centering
    \vspace{0.5cm}
    \hspace{0.35cm}
    \begin{minipage}[t]{0.45\linewidth}
        \centering
        \includegraphics[width=\linewidth]{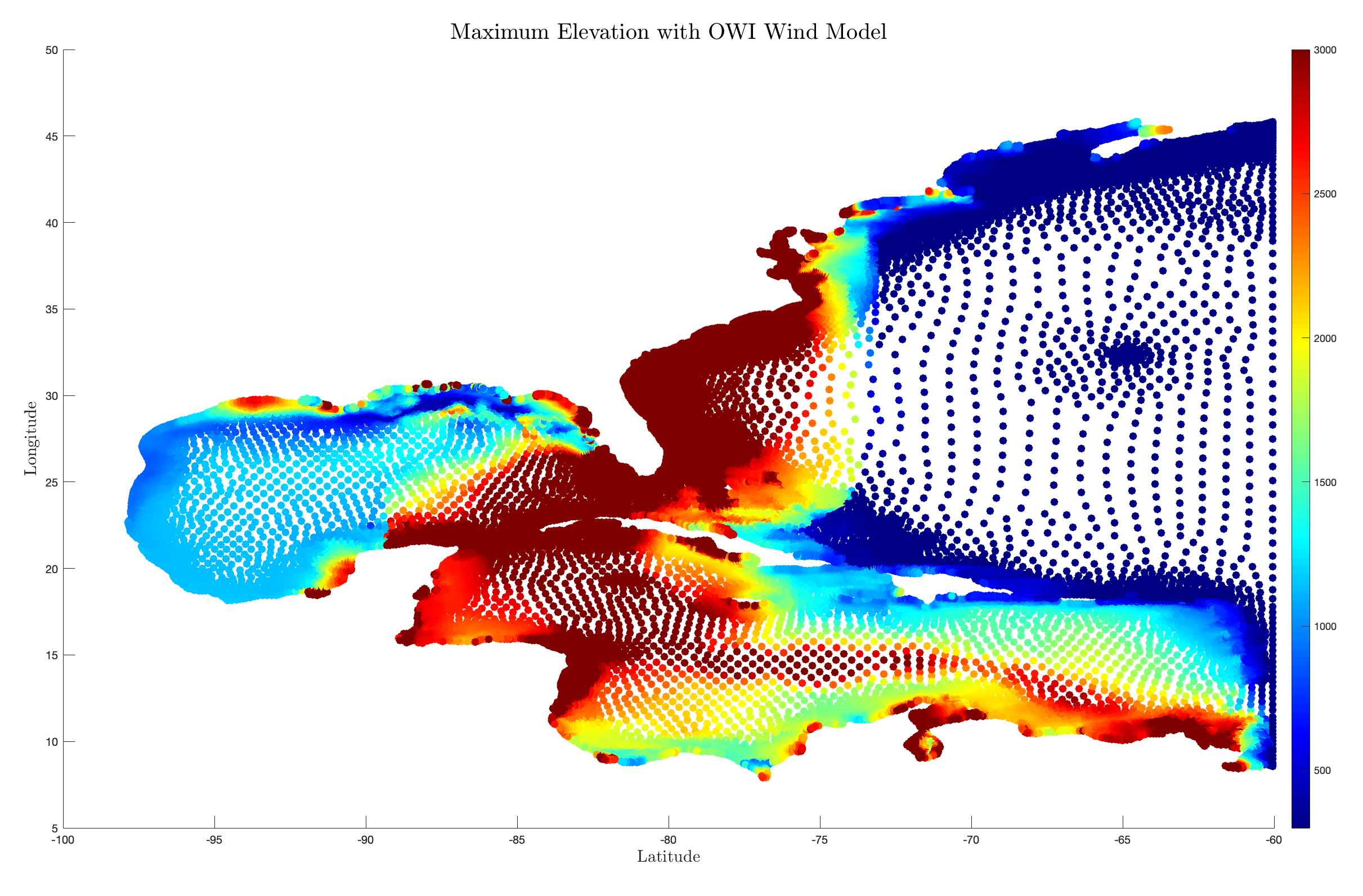}
    \end{minipage}%
    \hspace{-0.25cm}
    \begin{minipage}[t]{0.45\linewidth}
        \centering
        \includegraphics[width=\linewidth]{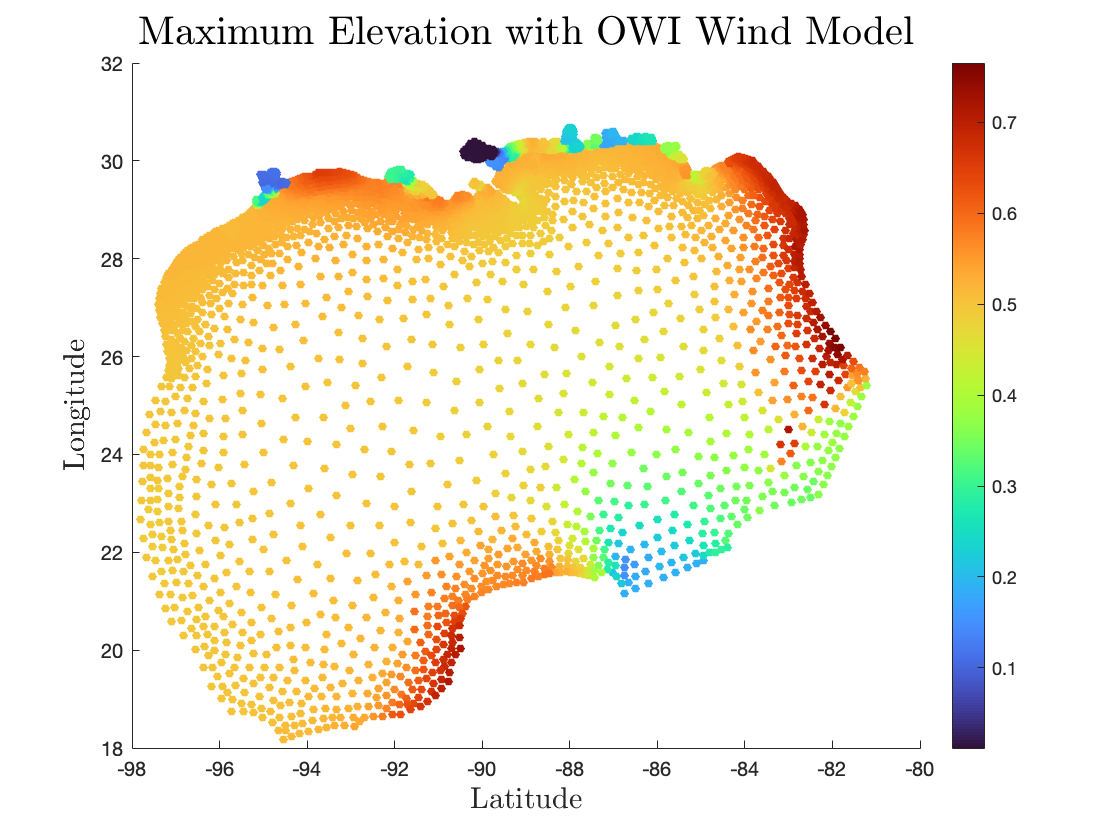}
    \end{minipage}
    \caption{Storm surge simulations illustrating unstructured data challenge.~(a) High-resolution ADCIRC mesh (31,000 nodes) showing maximum water levels (meters) for Gulf of Mexico region.~(b) Coarse mesh (8,000 nodes) for Florida region. Both use OWI wind forcing for Hurricane Ian (2022). Variable mesh dimensions ($d = 8,000$ vs. $31,000$) preclude standard fixed-size NPE methods.}%
    \label{fig:meshes}
\end{figure}

Flow-based NPE requires fixed-dimensional representations and is therefore sensitive to padding/truncation choices and to changes in geometry or resolution.
More broadly, amortized SBI models trained on a single representation often fail to generalize across meshes, sensor layouts, or measurement operators without retraining~\cite{gloeckler_all--one_2024}.
For instance, figure~\ref{fig:meshes} shows two different meshes for storm surge simulations, with varying node counts and spatial resolution. 
A flow-based NPE trained on one mesh would not be directly applicable to the other without retraining or interpolation, which can introduce additional errors.
These issues motivate architectures with permutation-invariant or operator-aware encoders, and, in some settings, diffusion models defined directly on function spaces to reduce discretization dependence~\cite{baldassari_conditional_2023}.

In summary, SBI's mathematical limitations stem from its foundational assumptions and algorithmic designs, which are often challenged by the complexities of real-world data that is incomplete, deviates from the assumed model, or exhibits high and varying dimensionality.

\subsection{Computational costs and trade-offs}\label{sect:comp_cost_tradeoffs}

Diffusion-based SBI is not a panacea. 
Improving robustness to the above data challenges typically increases computational cost.
We group the main trade-offs as follows:
\begin{itemize}
    \item \textbf{Iterative sampling overhead.} Posterior sampling often requires many denoising steps ($T\approx 10$--$1000$), whereas flow-based NPE can sample in one pass.
    Faster samplers and reduced-step methods can lower $T$ but introduce additional design complexity~\cite{karras2022elucidatingdesignspacediffusionbased, schmitt2024consistencymodelsscalablefast}.
    \item \textbf{Amortization gaps.} Some sequential diffusion-based SBI methods retrain or refine models for each observation, improving simulation efficiency but reducing amortization~\cite{sharrock2024sequentialneuralscoreestimation}.
    Composition-based methods trade flexibility for accumulated approximation error when aggregating evidence across observations~\cite{geffner2023compositionalscoremodelingsimulationbased}.
    \item \textbf{Hyperparameter sensitivity.} Noise schedules, solver choices, network architectures, and the diffusion horizon $T$ can be problem dependent and materially affect calibration and runtime.
\end{itemize}
As a segue to the forthcoming literature review, we establish evaluation criteria relevant to these data challenges and computational trade-offs.
We evaluate methods according to (i) which data challenges they target, (ii) computational costs (simulation budget and sampling time), and (iii) empirical performance and failure modes on benchmarks designed to expose missingness, irregular structure, and misspecification.

\section{Diffusion architectures for SBI:\ mechanisms and computational details}\label{sect:architectures}
Score-based diffusion models learn a reverse-time generative process from a fixed forward noising procedure via score matching (section~\ref{sect:diffusion_models}).
This process suffers from slow sampling speed and struggles with missing data imputation and data irregularity~\cite{song_generative_2020,song_score-based_2021}.
In SBI, one typically requires \emph{conditional} generation (e.g., sampling $\bftheta$ given observations $\bfx_{\mathrm{o}}$), robustness to missing or irregular inputs, and mechanisms to incorporate inference-time constraints or prior shifts (section~\ref{sect:data_limitations}).
This section summarizes five architectural strategies used in diffusion-based SBI and highlights the computational implications that recur across methods.

\subsection{Conditional diffusion models}
The default route to posterior inference with diffusion models is to learn a \emph{conditional score} for the noised parameter:
\begin{equation}\label{eq:cond_score_net}
s_{\phi}(\bftheta_t, t, \bfx_{\mathrm{o}})
\approx
\nabla_{\bftheta}\log p_t(\bftheta_t \mid \bfx_{\mathrm{o}}).
\end{equation}
Training data are generated from the simulator joint distribution
$\bftheta \sim p(\bftheta)$ and $\bfx\sim p(\bfx\mid\bftheta)$, and the model is trained with conditional denoising score matching.
In a DDPM-style parameterization, one may write $\bftheta_t=\sqrt{\bar{\alpha}_t}\,\bftheta_0+\sqrt{1-\bar{\alpha}_t}\,\boldsymbol{\varepsilon}$ with $\boldsymbol{\varepsilon}\sim\mathcal{N}(0,I)$, yielding the objective
\begin{equation}\label{eq:cond_DSM_loss}
\mathcal{L}(\phi)
= \mathbb{E}_{(\bftheta_0,\bfx_{\mathrm{o}}),\,t,\,\boldsymbol{\varepsilon}}
\left[
\left\|
s_{\phi}(\bftheta_t, t, \bfx_{\mathrm{o}})
- \nabla_{\bftheta}\log p_t(\bftheta_t \mid \bftheta_0)
\right\|_2^2
\right].
\end{equation}
The main design choice is how $\bfx_{\mathrm{o}}$ enters the score network.
Common strategies include 
\begin{itemize}
    \item concatenation of $(\bftheta_t,\bfx_{\mathrm{o}})$ for small, fixed-dimensional observations;
    \item attention or cross-attention for structured or variable-size observations;
    \item summary-network conditioning $\bfx_{\mathrm{o}}\mapsto h(\bfx_{\mathrm{o}})$ when $\dim(\bfx_{\mathrm{o}})$ is large or irregular.
\end{itemize}
Conditional diffusion is the workhorse architecture underlying posterior score estimation and conditional inverse-problem formulations used throughout the SBI diffusion literature (e.g.,~\citealp{tashiro_csdi_2021, chen_conditional_2025, baldassari_conditional_2023, nautiyal_condisim_2025, sharrock2024sequentialneuralscoreestimation, gloeckler_all--one_2024}).

~\cite{tashiro_csdi_2021}'s CSDI has been regarded as a foundational example of conditional diffusion models being successfully applied to generative tasks related to conditioning and inverse problems~\cite{simons_neural_2023,geffner2023compositionalscoremodelingsimulationbased,gloeckler_all--one_2024}.
CSDI, while not directly applied to SBI contexts, is explicitly trained for probabilistic time series imputation and forecasting, handling datasets with high missing ratios (e.g., 80\%). 
CSDI is a key advancement in using conditional diffusion for missing data problems, where training partitions the observed data into conditional information and imputation targets, enabling learning without access to true missing values.

\subsection{Guided diffusion models}\label{sect:guided_diffusion}
Guided diffusion has the capability to override the pre-programmed diffusion model map, particularly helpful in cases of model misspecification.
At each step of the reverse diffusion, external instructions such as new priors or non-linear constraints are injected to nudge the sampling trajectory toward desired characteristics, such as matching observed data or satisfying constraints~\cite{gloeckler_all--one_2024}.
This allows these models to deal with unexpected out-of-distribution scenarios without rebuilding the entire model.

A common form decomposes the target score as
\[
\nabla_{\bftheta}\log p_{\mathrm{target}}(\bftheta_t\mid\bfx_{\mathrm{o}})
=
\nabla_{\bftheta}\log p_{\mathrm{learned}}(\bftheta_t\mid\bfx_{\mathrm{o}})
+
\nabla_{\bftheta}\log g(\bftheta_t),
\]
where $p_{\mathrm{learned}}$ is the pre-trained model and $g$ encodes an analytic correction or constraint.

Examples include \emph{prior guidance} (reweighting from $p(\bftheta)$ to $q(\bftheta)$ via $g(\bftheta)\propto q(\bftheta)/p(\bftheta)$, requiring tractable prior scores) and \emph{constraint/energy guidance} that penalizes infeasible regions through differentiable penalties.
Guidance is especially attractive when simulations are expensive, since it can reuse a trained score model without retraining, but it can fail when the guided target lies outside the support represented by the learned model.
Guided diffusion is widely used for inverse problems and has been adapted to SBI settings (e.g.,~\cite{song_pseudoinverse-guided_2022, chang_inference-time_nodate, gloeckler_all--one_2024}).

\subsection{Sequential diffusion models}
Sequential diffusion-based SBI improves simulation efficiency for a fixed observation by iteratively refining the training distribution toward posterior-relevant regions.
Across rounds $r=1,\dots,R$, one simulates from a proposal $q^{(r)}(\bftheta)$, trains (or updates) a conditional score model on the resulting pairs, and updates the proposal using the current posterior approximation.

This inherits the core benefits of sequential SBI (e.g., focusing simulations where they matter) while retaining score-based posterior sampling in the final round; SNPSE and variants follow this template~\cite{sharrock2024sequentialneuralscoreestimation}.
The main trade-off is a reduced degree of amortization, since the procedure is tailored to one (or a small set of) observations.

\subsection{Compositional/factorized diffusion}
Compositional or factorized architectures target settings with multiple observations by enabling inference-time aggregation.
A representative approach is factorized neural posterior score estimation (F-NPSE), which exploits additivity of log-densities under conditional independence assumptions to compose information from multiple observations using a single trained score model (up to prior corrections)~\cite{geffner2023compositionalscoremodelingsimulationbased, liu_compositional_2023}.
These methods can reduce simulator calls per observation and support flexible evidence aggregation, but accuracy depends on the validity of the factorization assumptions and on stability of score composition.

\subsection{Consistency models}\label{sect:consistency_models}
Consistency models (CMs) are a class of diffusion-based models that utilize a  Consistency Model Posterior Estimation (CMPE) methodology~\cite{kelly_simulation-based_2025,schmitt2024consistencymodelsscalablefast}.
Consistency models address this by learning a map that sends noisy states back to a clean sample, enabling few-step or one-step generation.
Rather than estimating a score, a consistency model learns $f_\phi(\bftheta_t,t)\approx \bftheta_0$ subject to a trajectory-consistency constraint (all points on the same probability-flow trajectory map to the same output).

In SBI, CMPE uses this idea to produce fast conditional posterior samples while retaining the flexibility of unconstrained neural architectures~\cite{schmitt2024consistencymodelsscalablefast}.
The trade-off is that performance depends on the quality of the underlying diffusion/flow being distilled and the stability of the distillation objective.

\noindent \textbf{When not to use diffusion-based SBI.}
Diffusion-based SBI is most compelling when likelihoods are intractable, posteriors are complex, and observations are irregular or high-dimensional.
However, alternatives can be preferable in common regimes:
\begin{itemize}
    \item if $\dim(\bftheta)$ is small and likelihood evaluation is feasible, MCMC may be favored for asymptotic exactness;
    \item if the posterior is approximately unimodal and well-approximated locally, Laplace or variational approximations can be effective;
    \item if the simulator is differentiable and gradients are reliable, gradient-based Bayesian methods (e.g., HMC) may be competitive;
    \item if real-time inference is required and diffusion sampling is too slow, distillation or flow-based amortization may be preferable.
\end{itemize}
Table~\ref{tab:architectures} categorizes the surveyed methods by architecture and data challenges addressed, providing a high-level overview of the landscape of diffusion-based SBI approaches.
For further details on each architecture, there are several tables included in appendix~\ref{app:comp_complexity} that summarize the computational complexity, implementation guidelines, and failure modes of each method.
Table~\ref{tab:arch_selection} provides guidance on when to use each
method, table~\ref{tab:pract_imp} lists practical implementation details,
and table~\ref{tab:failure-modes} summarizes common failure modes and
mitigations. Computational costs for each architecture are summarized in
table~\ref{tab:comp_complexity}.

In addition to these core  architectures, there are related models that are worth mentioning for their contributions to relevant problems with challenging data.
Among these are neural SDMs~\cite{bartosh2024neuraldiffusionmodels}, DSB/CDSB (Schrödinger Bridge)~\cite{debortoli2023diffusionschrodingerbridgeapplications,shi_conditional_2022}, structure diffusion (GSDM)~\cite{weilbach_graphically_2023}, and multi-speed Diffusion~\cite{batzolis_conditional_2021}.

Now, we have established the algorithmic and computational foundations for understanding the methods surveyed in the next section, where we evaluate how these architectures address the data challenges of section~\ref{sect:data_limitations} in practice.

%%% TABLE 1: Method categorization
\begin{table}[t]
\centering
\caption{Categorization of diffusion-based SBI methods and corresponding
data limitations addressed.}
\label{tab:architectures}
\begin{tabular*}{\textwidth}{@{\extracolsep{\fill}}
  p{3cm} p{2.5cm} p{5.5cm}}
\hline
\textbf{Model type} & \textbf{Method} & \textbf{Data issues addressed} \\
\hline
\multirow{3}{3cm}{Conditional diffusion}
  & Cond.\ SDMs~\cite{baldassari_conditional_2023}
  & Irregular data (format): inverse problems in infinite-dimensional
    function spaces; discretization-invariant inference \\
  & cDiff~\cite{chen_conditional_2025}
  & Irregular data (format): varying sequence lengths; training
    instability of normalizing flows \\
  & ConDiSim~\cite{nautiyal_condisim_2025}
  & Irregular data (complexity): high-dimensional, multimodal
    posteriors with intractable likelihoods \\
\hline
\multirow{2}{3cm}{Sequential diffusion}
  & SNPSE~\cite{sharrock2024sequentialneuralscoreestimation}
  & Standard data \\
  & SeqDiff~\cite{stevens_sequential_2025}
  & Irregular data (format): sequential/video data; slow sampling
    via temporal structure exploitation \\
\hline
\multirow{2}{3cm}{Guided diffusion}
  & PriorGuide~\cite{chang_inference-time_nodate}
  & Model misspecification/OOD: inference-time prior adaptation
    without retraining \\
  & Simformer~\cite{gloeckler_all--one_2024}
  & Irregular data, model misspecification, and missing data \\
\hline
Factorized diffusion
  & F-NPSE~\cite{geffner2023compositionalscoremodelingsimulationbased}
  & Standard data \\
\hline
Consistency models
  & CMPE~\cite{schmitt2024consistencymodelsscalablefast}
  & Standard data (with noise and distractors) \\
\hline
\end{tabular*}
\end{table}

\section{Survey of existing works}\label{sect:lit_survey}
Recent work has established diffusion models as a flexible alternative to flow-based and likelihood-based approaches for simulation-based inference (SBI).
Within this literature, diffusion-based SBI methods differ primarily in how conditioning is incorporated, how simulation efficiency is achieved, and whether robustness to data irregularity or model misspecification is explicitly addressed.

We identify eight representative diffusion-based SBI methods.
Three focus on {standard SBI settings} with fixed-dimensional, complete data, while the remaining methods address at least one of the three data challenges introduced in section~\ref{sect:data_limitations}: unstructured data, missing or corrupted observations, and model misspecification.
Table~\ref{tab:lit} summarizes how these methods align with these challenges.
Below, we group the literature by the dominant data regime rather than by algorithmic chronology.

%%% TABLE: Literature summary
\begin{table}[t]
\centering
\caption{Diffusion-based SBI methods and the data limitations they address.
The works of~\cite{sharrock2024sequentialneuralscoreestimation,
geffner2023compositionalscoremodelingsimulationbased,
schmitt2024consistencymodelsscalablefast} are omitted as they are
demonstrated only on standard data.}
\label{tab:lit}
\begin{tabular*}{\textwidth}{@{\extracolsep{\fill}} cccc}
\hline
\textbf{Paper}
  & \textbf{Missing data}
  & \textbf{Unstructured data}
  & \textbf{Model misspecification} \\
\hline
\cite{baldassari_conditional_2023}
  &
  & \checkmark
  & \\
\cite{gloeckler_all--one_2024}
  & \checkmark
  & \checkmark
  & \checkmark \\
\cite{chang_inference-time_nodate}
  &
  &
  & \checkmark \\
\cite{chen_conditional_2025}
  &
  & \checkmark
  & \\
\cite{nautiyal_condisim_2025}
  &
  & \checkmark
  & \checkmark \\
\hline
\end{tabular*}
\end{table}

\subsection{Standard data}\label{sect:standard_data}
Several works adapt diffusion models to SBI under the assumption of well-structured, fixed-dimensional, and complete observations.
Such frameworks offer promise, but due to their focus on standard data settings, they may not directly address the challenges posed by complex data found in real-world applications.
However, they lay the groundwork for future adaptations to more challenging data scenarios and provide insights into the integration of diffusion models with SBI.\
These methods primarily aim to improve simulation efficiency, posterior expressivity, or sampling speed relative to normalizing-flow-based approaches.
Table~\ref{tab:standard_data} summarizes these methods, highlighting their key features and contributions.

\subsubsection{Sequential Neural Posterior Score Estimation (SNPSE)}\label{sect:snpse}

Sharrock et al.~\cite{sharrock2024sequentialneuralscoreestimation} introduce SNPSE, which applies conditional score-based diffusion models to neural posterior estimation (NPE) in a sequential SBI framework.
The motivation behind SNPSE is particularly aimed at improving efficiency and robustness in standard SBI settings with well-structured data.

Rather than learning a globally amortized posterior, SNPSE iteratively refines a proposal distribution toward a fixed observation, reducing simulation cost by adaptively concentrating simulations around regions of high posterior mass.
This cost saving is due to sequential component; each iteration samples new parameters from the previously estimated posterior rather than from the prior.
SNPSE avoids the normalization constraints and support mismatch issues of flow-based NPE, but sacrifices amortization and assumes complete, fixed-dimensional observations.

Across benchmark tasks---including two-moons, Gaussian mixtures, and a real-world neuroscience application---SNPSE demonstrates competitive or superior accuracy to SNPE and SNLE.\
These results suggest that sequential posterior refinement can substantially improve robustness in standard, well-structured inference problems.

Geffner et al.~\cite{geffner2023compositionalscoremodelingsimulationbased}
(cf.~section~\ref{sect:fnpse}) pursue a related line but emphasize
global amortized NPSE rather than sequential refinement, trading
per-observation efficiency for broader amortization across observations
Additionally, SNPSE integrates the ideas of PriorGuide~\cite{chang_inference-time_nodate}, a method specifically developed to resolve model misspecification issues.

While only applied in standard data contexts in this work, SNPSE offers potential for mitigating prior-data mismatch and improving performance under broad or mildly misspecified priors.
In contrast to inference-time guidance methods such as~\cite{chang_inference-time_nodate} (cf.~section~\ref{sect:priorguide}), which adjust a pre-trained prior model only at sampling time, SNPSE integrates this adaptation directly into training.
The adaptive sampling inherent to sequential NPSE helps refine the exploration to regions most relevant to the observed data, offering a pathway to addressing scenarios where a broad initial prior might be ``misspecified'' for the specific observation.

\subsubsection{Factorized Neural Posterior Score Estimation (F-NPSE)}\label{sect:fnpse}
Geffner et al.~\cite{geffner2023compositionalscoremodelingsimulationbased} propose F-NPSE, which exploits conditional independence across observations to factorize posterior score estimation.
F-NPSE's core contribution is balancing simulation efficiency with accuracy, addressing key shortcomings of existing neural SBI methods NPE, NLE, and NRE.\

By training only on single observation–parameter pairs and aggregating scores at inference time, F-NPSE achieves substantial simulation efficiency gains for multi-observation inference.
This factorization trades off accuracy for efficiency and relies on independence assumptions, but avoids the MCMC requirement associated with traditional neural methods and remains robust to multimodal posteriors.

A disadvantage of F-NPSE is that aggregating scores derived from approximations (factorized inference) can accumulate error, whereas a fully conditional NPSE (non-factorized, trained on all $n$ observations) or NPE avoids this aggregation error but sacrifices simulation efficiency.
This illustrates a key through-line comparison between F-NPSE and SNPSE~\cite{sharrock2024sequentialneuralscoreestimation}.

F-NPSE, in its core formulation, assumes that observations are independent and identically distributed (i.i.d.), but regardless, conceptual pathways exist for extending its application.
For example, F-NPSE could be combined with sequencing or imputation mechanisms found in works such as~\cite{tashiro_csdi_2021,geffner2023compositionalscoremodelingsimulationbased, verma_robust_2025,sharrock2024sequentialneuralscoreestimation}.
Such structural encoding could enable the method to account for observations that vary in both format and length.

In summary, F-NPSE represents a significant step forward in SBI by leveraging factorization to improve simulation efficiency while maintaining the flexibility of conditional diffusion models.
It shows promise for future adaptations to non-ideal data scenarios with space to incorporate sequential refinement strategies like those in SNPSE~\cite{sharrock2024sequentialneuralscoreestimation} or compositional approaches like those in~\cite{geffner2023compositionalscoremodelingsimulationbased}.

\subsubsection{Consistency Models for Scalable and Fast Simulation-Based Inference (CMPE)}\label{sect:cmpe}
~\cite{schmitt2024consistencymodelsscalablefast} introduce {Consistency Model Posterior Estimation (CMPE)}, an SBI framework grounded in the theory of {consistency models}, (cf.~section~\ref{sect:architectures}). 
Unlike conventional diffusion-based SBI methods, which rely on iterative score refinement across noise levels, CMPE trains a diffusion-based parametric map that enforces a consistency property between posterior distributions at different noise levels.

CMPE avoids the pitfalls of additive score composition approaches by construction, which~\cite{geffner2023compositionalscoremodelingsimulationbased} critique as mathematically inconsistent for posterior inference, specifically since the assumptions of additive score composition fail unless restrictive independence conditions hold.
Moreover, CMPE requires only a small number of model evaluations (e.g., 2--4 steps) to generate samples from the approximate posterior.
This is an advantage in comparison to older methods such as NPSE or NLSE~\cite{simons_neural_2023}, which require many iterative denoising steps.

CMPE does not yet address non-ideal data challenges~\cite{simons_simulation-based_nodate}.
The scope of this material remains in the class of standard synthetic SBI tasks, leaving open questions about its robustness under more realistic data regimes.
The data benchmarks, while standard, are designed to include noise and distractors, which are common in real-world applications, thus providing a more realistic testing ground for the method's performance.

Given that consistency models have been successfully applied to unstructured modalities such as images, audio, and video in general generative modeling~\cite{schmitt2024consistencymodelsscalablefast}, CMPE could in principle be extended to handle unstructured data in SBI, potentially combining the few-step efficiency of consistency-based inference with the flexibility of diffusion-based methods.
Future work could explore CMPE's adaptation to irregularly sampled time series, spatial fields, or function-valued parameters, making use of the inherent adaptability of consistency models to diverse data formats.

\medskip

\noindent Together, these methods demonstrate that diffusion models can match or outperform flow-based SBI in standard settings, motivating their extension to more realistic data regimes.

%%% TABLE: Standard data methods comparison
\begin{table}[t]
\centering
\caption{Comparison of model type, core contribution, and relation to
neural SBI for the three standard-data methods of
section~\ref{sect:standard_data}.}
\label{tab:standard_data}
\begin{tabularx}{\textwidth}{l X X X}
\hline
\textbf{Feature} & \textbf{CMPE} & \textbf{F-NPSE} & \textbf{SNPSE} \\
\hline
Model type
  & Consistency models
  & Conditional SDMs
  & Conditional SDMs \\
\hline
Core contribution
  & Fast few-step inference retaining diffusion flexibility;
    $100\text{--}1000\times$ faster than continuous-flow methods
  & Simulation efficiency for multiple observations via score
    factorization; one simulator call per training case
  & Sequential proposal refinement for a single observation;
    adaptively guides simulations toward high-posterior regions \\
\hline
Neural SBI comparison
  & Competitive, unconstrained alternative to normalizing flows;
    resolves slow sampling of continuous-flow methods for amortized
    inference
  & Addresses simulation inefficiency of NPE for sets of observations;
    annealing sampler robust to multimodality avoids MCMC requirement
    of NLE/NRE
  & Achieves comparable or superior accuracy to sequential NPE (SNPE);
    avoids architectural restrictions of flow-based approaches \\
\hline
\end{tabularx}
\end{table}

\subsection{Unstructured data}\label{sect:unstruct_lit}
Real-world scientific data often violate the assumptions of fixed dimensionality or complete observation.
When encountered with data of mismatched dimensions, for instance when observations are high-dimensional but the parameters are low-dimensional, the model spends most of its capacity processing $\bfx_\text{o}$ rather than learning the posterior over $\bftheta$.
This is due in large part to a lack of summary mechanisms that can distill irregular inputs into compact representations that retain relevant information for inference~\cite{chen_conditional_2025}.
Diffusion-based SBI methods that can provide relevant solutions to accommodate such unstructured data are crucial for broadening applicability to practical scientific problems.

We have included three papers below; one which addresses cases of function-valued parameters and infinite-dimensional settings~\cite{baldassari_conditional_2023}, one which focuses on irregularly sampled time series and data of varying lengths and dimensions~\cite{chen_conditional_2025}, and one that~\cite{nautiyal_condisim_2025} considers noisy, high-dimensional data with distractors, which can be viewed as a form of unstructured data complexity.

Before proceeding, we distinguish the relevance of these methods to missing data scenarios, as in some cases, ``missingness'' can be interpreted as a form of unstructured data.
A direct comparison is difficult because the source materials do not formally describe methods for dealing with traditional missing data imputation or masking schemes (like those used in Neural Posterior Estimation (NPE) variants such as~\cite{verma_robust_2025,wang_missing_2024} and CSDI~\cite{tashiro_csdi_2021}). 
Instead, they focus on higher-level types of unstructured data complexity.
However, these forms of unstructured complexity often implicitly address challenges similar to those encountered in missing data scenarios.

In Baldassari's work,``missingness'' is manifested as the ill-posed nature of the inverse problem, where observations are noisy or sparse.
Condisim provides an implicit mechanism that could bridge to handling missing data by isolating informative parameters from distractors in data that is ``heavily corrupted''. 
Although there are not distinct works that relate diffusion-based SBI directly to standard missing data imputation, the unstructured data frameworks discussed below have the potential to inherently tackle similar challenges of data incompleteness and irregularity.

\subsubsection{Conditional SDMs: addressing infinite dimensions}\label{sect:baldassari}
Baldassari et al.~\cite{baldassari_conditional_2023} provide the first rigorous formulation of conditional SDMs in infinite-dimensional Hilbert spaces.
Motivated by~\cite{Stuart_2010}'s principle of ``avoiding discretization until the last possible moment'', their framework achieves discretization-invariant Bayesian inference for function-valued parameters, avoiding grid-dependent artifacts that arise when naively discretizing PDE-based inverse problems.
This work establishes foundational theoretical guarantees for conditional diffusion in infinite dimensions using a proof from~\cite{batzolis_conditional_2021} and the principles of Pidstrigach formalism~\cite{pidstrigach2025infinitedimensionaldiffusionmodels}, making it particularly relevant for scientific applications with functional unknowns.

While there exists previous attempts to project function-valued unknowns onto finite-dimensional representations for diffusion modeling~\cite{DupontFuncta, phillips2022spectraldiffusionprocesses}, or to generalize diffusion to function spaces without fully developing the time-continuous SDE limit~\cite{kerrigan2023diffusiongenerativemodelsinfinite}, these approaches are not discretization-invariant, and posterior inferences can change as the grid is refined.

Considering these research gaps, the authors present four crucial ideas: a formal definition of the conditional score for the infinite-dimensional case, theoretical guarantees for convergence of the reverse SDE under Gaussian and more general priors, conditions ensuring stability and valid sampling from the target conditional distribution, and a proof that the conditional score can be consistently estimated using conditional denoising score matching. 
This avoids the inconsistencies highlighted by~\cite{geffner2023compositionalscoremodelingsimulationbased} and related work~\cite{liu_compositional_2023} that rely on additive score factorization.

Through examples, the authors demonstrate that their method supports large-scale, discretization-invariant Bayesian inference with strong theoretical foundations and practical efficiency.
The method is likelihood-free, amortized, and tailored to function-valued parameters, making it particularly relevant for SBI in PDE-driven applications and other settings where the unknowns live in function spaces rather than finite-dimensional vectors.

\subsubsection{cDiff: addressing diverse data structures}
Chen et al.~\cite{chen_conditional_2025} introduce cDiff, a conditional diffusion model for amortized NPE that incorporates learned summary networks.
By mapping variable-sized datasets (e.g., sets or sequences) into fixed-dimensional embeddings, cDiff enables stable posterior inference for irregular and sequential data.

The authors note contrasts between cDiff and existing NF techniques.
Flow-based NPE is knowingly limited by the need for architectures that assume fixed-length or grid-structured data; diffusion models without summary networks similarly struggle and can lead to amortized models containing instability, large variance in score estimates, and inefficiency when the observation dimension is large but the parameter dimension is small.
Chen et al.\ address this challenge by jointly training a summary encoder and a conditional diffusion decoder, building on the preceding work of SNPSE~\cite{sharrock2024sequentialneuralscoreestimation}, F-NPSE~\cite{geffner2023compositionalscoremodelingsimulationbased}, and Flow Matching Posterior Estimation~\cite{dax_flow_2023}.
Functionally, the summary network $s_\psi(\cdot)$ compresses datasets of any size into a fixed-dimensional embedding, while the diffusion decoder $q_\phi(\bftheta\mid s_\psi(\bfx))$ models the posterior over parameters, along with key theoretical justifications.
This capability handles variability in the amount of data observed (length of sequence).

The authors demonstrate cDiff on a benchmark suite spanning three categories---no-encoder tasks (fixed-size inputs), IID variable-size datasets, and
sequential variable-length datasets encoded with biLSTMs~\cite{radev_bayesflow_2022}.
Across nearly all tasks, conditional diffusions outperform normalizing flows in accuracy, calibration, and stability, while training substantially faster.

The results demonstrate that diffusion models coupled with learned summary networks offer a robust, amortized solution for SBI with datasets whose sizes or structures vary across simulations.
In summary, like other NPE methods, this method remains vulnerable to model misspecification and out-of-distribution inputs, which can degrade posterior reliability when applied to unstructured or real-world data. 
Nonetheless, this work provides an important step toward SBI methods that generalize across diverse, irregular, and high-dimensional data modalities.

\subsubsection{ConDiSim: addressing noise and distraction}\label{sect:condisim}
Nautiyal et al.~\cite{nautiyal_condisim_2025} propose ConDiSim, an amortized conditional diffusion model designed to operate directly on raw, noisy, or distractor-heavy observations.
Unlike methods that rely on handcrafted summaries, ConDiSim suppresses irrelevant observation dimensions through the learned conditional denoising process.

Conceptually, ConDiSim is closely related to other diffusion-based SBI approaches that learn conditional posterior estimators, such as~\cite{chang_inference-time_nodate} and~\cite{sharrock2024sequentialneuralscoreestimation}. 
Unlike sequential methods (e.g., SNPSE/TSNPSE), however, ConDiSim does not refine per-observation but amortizes across $\xobs$ via a conditional reverse diffusion on parameters.

ConDiSim allows for efficient posterior evaluation but entails careful choices of hyperparameters to ensure stability and high-quality samples. 
Sampling requires iterative reverse steps (as in DDPMs), which can become costlier than non-iterative SBI methods.
Despite this, across SBI benchmarks with heavy observational corruption, ConDiSim maintains calibrated posteriors, highlighting diffusion models' robustness to unstructured data complexity, notably in comparison to existing approaches such as GATSBI~\cite{ramesh_gatsbi_2021}, SNPE~\cite{sharrock2024sequentialneuralscoreestimation}, NPE~\cite{greenberg_automatic_2019}, and Simformer~\cite{gloeckler_all--one_2024}.

ConDiSim brings forth questions regarding model misspecification, as the noisy observation model can differ from the idealized simulator used for training.  
While ConDiSim does not target model misspecification as conventionally defined in the literature, the use of distractors has the potential to tackle scenarios where observation data poorly represents the training data.

Altogether, ConDiSim's conditional denoising formulation, robustness to irrelevant dimensions, and amortized posterior modeling make it a practical and flexible tool for modern SBI settings where observations are high-dimensional, imperfect, or lacking clear sufficient statistics.

\medskip

\noindent 
In summary, these works demonstrate that conditional diffusion models can naturally accommodate irregular inputs, function-valued parameters, and high-dimensional observations without restrictive architectural constraints.

\subsection{Model misspecification}
Model misspecification remains a central challenge in SBI, particularly when priors or simulators used during training differ from those relevant at inference time.
When new prior knowledge becomes available at runtime, typically, the entire score model must be retrained, which can be expensive and inefficient~\cite{chang_inference-time_nodate}. 

We find one work, PriorGuide~\cite{chang_inference-time_nodate, yang2025priorguidetesttimeprioradaptation}, that addresses model misspecification for diffusion-based SBI.\
Even so, the intersection of diffusion-based SBI and model misspecification remains an underexplored and active area, with many open questions regarding robustness to OOD data and structural model flaws~\cite{kelly_simulation-based_2025}.

\subsubsection{Inference-Time Prior Adaptation via Guided Diffusion Models (PriorGuide)}\label{sect:priorguide}
Chang et al.~\cite{chang_inference-time_nodate,yang2025priorguidetesttimeprioradaptation} introduce PriorGuide, a guided diffusion framework that enables test-time prior adaptation without retraining.
By decomposing the posterior score into a learned component and an analytic prior correction, PriorGuide steers a pre-trained diffusion model toward a new prior distribution during sampling, addressing two related sources of misspecification:
\begin{enumerate}
    \item prior mismatch, where the training prior differs from the true parameter distribution, and
    \item out-of-distribution observations, where $\xobs$ lies far outside the region supported by the training prior.
\end{enumerate}

The method differs from related work on prior specification from~\cite{elsemuller_sensitivity-aware_2024, chang2025amortizedprobabilisticconditioningoptimization,whittle2025distributiontransformersfastapproximate}, which generally do not support arbitrary test-time priors.

Rather than regenerating simulations or retraining, PriorGuide rewrites the target posterior $q(\bftheta\mid\xobs)$ as a reweighted version of the original posterior $p(\bftheta\mid\xobs)$ and includes it as an additional guidance term for the reverse diffusion process.
At each diffusion time step $t$, the posterior score decomposes into the learned score from the base model and a correction based on the prior ratio $\rho(\bftheta)=q(\bftheta)/p(\bftheta)$.
This yields a simple mean-shift update that integrates new prior information while preserving the existing amortized posterior estimator.

Using Simformer~\cite{gloeckler_all--one_2024} as the base diffusion model, PriorGuide demonstrates that guided sampling is competitive with retraining under new priors on standard SBI benchmarks.
Practically, the method is most beneficial in settings where the training prior is only an approximation—often chosen for computational convenience, regenerating simulations under a corrected prior is prohibitively expensive. 

This method does not come without limitations.
PriorGuide requires access to the likelihood score, which can be expensive in high dimensions, and posterior adaptation depends on the expressivity of the pre-trained diffusion model; if the model poorly covers the support of the true prior, then the guided correction cannot fully recover the correct posterior.
Also, as with other diffusion approaches, stability carefully depends on discretization and hyperparameter choices.

Overall, PriorGuide provides a lightweight mechanism for adapting posteriors when the true prior is narrower, heavier-tailed, or shifted relative to the prior used during training.
This makes guided diffusion a promising tool for robust SBI when simulation budgets are limited and true priors evolve or differ from those assumed during training.

\subsection{``All-in-one'' Simulation Based Inference}\label{sect:simformer}
There is one method in the literature that claims to address the three salient data challenges in SBI:\ unstructured data, missing data, and misspecified priors.
Shortly predating PriorGuide~\cite{chang_inference-time_nodate}, Gloeckler et al.~\cite{gloeckler_all--one_2024} propose the Simformer, a transformer-based diffusion model trained on the joint distribution of parameters and data.

The Simformer's novel architecure combines probabilistic diffusion models and transformers~\cite{peebles_scalable_2023} to perform both NLE and NPE based on ideas from~\cite{weilbach_graphically_2023}.
By representing both observed and unobserved variables as tokens with condition indicators and identical dimensions, the Simformer supports inference with missing data, unstructured observations, and function-valued parameters within a single architecture.
Guided diffusion enables inference-time constraint enforcement and prior adaptation, while attention mechanisms encode domain-specific conditional independence structure.

To handle function-valued or infinite-dimensional data, the Simformer incorporates the continuous indices (e.g., time points) into the variable representation using random Fourier feature embeddings.
The approach of using neural processes and Fourier feature embeddings to handle continuous domains has been empirically validated as discretization-invariant~\cite{baldassari_conditional_2023}.
This makes the method applicable to domains where parameters change continuously, without needing to commit to a fixed, dense grid resolution beforehand.

In challenging ecological models (Lotka-Volterra), the Simformer provided consistent inference even when observations were unstructured and irregularly placed in time.
The Simformer successfully infers infinite-dimensional parameters, such as the time-dependent contact rate in the SIRD epidemiological model, and the resulting posterior estimates were well-calibrated, demonstrating applicability far beyond fixed, finite-dimensional parameter spaces.
Moreover, Simformer exemplifies success in a large-scale geophysics imaging problem requiring the estimation of $256\times256$ parameters.
However, sampling is slower than for normalizing flow–based methods but faster than MCMC based approaches, as the quadratic scaling of transformer evaluations with input length also poses significant memory and computational challenges, though sparsity and attention-masking strategies can mitigate this.

The Simformer has inspired further extensions to complex scientific domains.
For example, SpatFormer~\cite{tesso_spatformer_2025} branches directly off the Simformer, adapting it specifically for problems in spatial statistical modeling. 
It uses the underlying conditional diffusion and transformer architecture and modifies the input pipeline by designing a specific tokenizer that embeds spatial coordinates alongside data values into the transformer.
This method solves the tractability and scalability issue inherent in MCMC-based inference for spatial models with GP priors by replacing them with an amortized approach trained using the Simformer framework

In summary, the Simformer unifies conditional, guided, and amortized diffusion-based SBI within a single framework.
Empirical results demonstrate robustness to missingness, irregular sampling, and prior mismatch across a range of scientific models, at the cost of increased computational complexity relative to simpler diffusion architectures.

\subsection{Summary and positioning}
Collectively, existing diffusion-based SBI methods demonstrate clear advantages over flow-based approaches in expressivity, robustness, and architectural flexibility.
However, current methods typically address only subsets of the challenges posed by unstructured data and model misspecification.
Table~\ref{tab:lit} situates these works along the axes of data irregularity and robustness, motivating the need for diffusion-based SBI frameworks that integrate principled uncertainty quantification with resilience to real-world data complexity.

\section{Conclusions and Future Directions}\label{sect:conclusion}
This review surveyed eight recent diffusion-based SBI methods and organized them around three data challenges that most often compromise posterior reliability in practice: unstructured observations, missing or corrupted data, and model misspecification.
By linking these challenges to the mechanisms reviewed in section~\ref{sect:data_limitations}--\ref{sect:architectures}, we highlighted how conditional, guided, sequential, factorized, and consistency-based diffusion architectures trade off flexibility, robustness, and computational cost.
Among these approaches, there are notable intersections and complementarities, particularly in how they handle unstructured data and adapt to model misspecification.

Across the literature, diffusion-based SBI offers clear benefits over classical SBI and flow-based NPE in expressivity and architectural freedom.
Relevant research to diffusion-based SBI is quickly proliferating, and several open questions and future directions emerge from the existing literature and ideas discussed in this review:
\begin{itemize}
    \item Incorporating arbitrary priors at runtime remains an open goal~\cite{chang_inference-time_nodate}.
    \item Robustness under misspecification may benefit from optimal-transport-based calibration and broader objective choices (e.g., GVI)~\cite{kelly_simulation-based_2025}; diffusion and flow-matching families remain underexplored in this context~\cite{gloeckler_all--one_2024, simons_neural_2023}.
    \item Sequential variants beyond those studied here (e.g., FMPE) and more specialized architectures—akin to those optimized in other diffusion modalities—are promising directions~\cite{sharrock2024sequentialneuralscoreestimation,karras2022elucidatingdesignspacediffusionbased}
    \item Modeling systematics and ``unknown unknowns'' remains a pressing need in cosmological and related SBI pipelines~\cite{Wang__2023}.
    \item Benchmark breadth: active learning/BO, gray-box methods, and GP-integrated hybrids were not covered in some surveys~\cite{lueckmann_benchmarking_2021}, and tasks with high-dimensional spatial structure (e.g., images) demand algorithms that learn summaries while exploiting structure~\cite{lueckmann_benchmarking_2021}.
\end{itemize}

These directions are strongly motivated by geophysical uncertainty quantification, where parameters and observations are often function-valued (space/time), irregularly sampled, noisy, and incomplete.
In storm surge modeling with ADCIRC~\cite{adcirc2004}, for example, uncertainties in bathymetry, friction, and meteorological forcing can strongly affect predicted extremes~\cite{mayo_climate_2022}.
Posterior inference from observed water levels can narrow plausible parameter regimes, propagate uncertainty into surge scenarios, and support probabilistic decision-making for emergency management and long-term resilience planning.

This idea can extend to geophysical models that feed into ADCIRC such as wave models (e.g., SWAN~\cite{holthuijsen_wave_2007}) and atmospheric models (e.g., WRF~\cite{skamarock_wrf_2008}), which can propagate uncertainties through the simulation, necessitating thorough understanding of how input parameter uncertainties affect model outputs.
A major practical bottleneck is forcing uncertainty: high-fidelity wind fields (e.g., OWI) are rarely available in real time, and simplified parameterizations (e.g., Holland) introduce substantial mismatch between the simulator used for inference and the evolving physical event.

The methods reviewed here suggest complementary ingredients for such settings.
For instance, PriorGuide~\cite{chang_inference-time_nodate} provides a useful foundation for adapting to new prior information without retraining a new model, which is crucial when new data or expert knowledge becomes available.
This framework could potentially be adapted to real-time forecasting scenarios where prior distributions of uncertain parameters need to be updated dynamically as new observations are collected during an evolving weather event.

Baldassari et al.~\cite{baldassari_conditional_2023} provide a principled foundation for discretization-invariant inference when unknowns are fields.

In a similar vein, ConDiSim~\cite{nautiyal_condisim_2025} has the potential to play an important role in quantifying uncertainties in climate simulation models. 
Due to the inherent complexity of climate systems, observational data is often noisy, incomplete, or contains irrelevant features that do not inform the parameters of interest.

cDiff~\cite{chen_conditional_2025} offers amortized conditioning through learned summaries for variable-format observations, while ConDiSim~\cite{nautiyal_condisim_2025} demonstrates robustness to noise and distractors.
Simformer~\cite{gloeckler_all--one_2024} is the closest existing attempt to unify missingness, unstructuredness, and misspecification within a single architecture, albeit at increased computational cost.

Naturally, any of these eight novel methods could be combined or extended to further enhance their capabilities in addressing the complex data challenges inherent in uncertainty quantification.
A central takeaway is that practical diffusion-based SBI for scientific models will likely require {hybridization}---combining principled function-space formulations, robust conditioning/summary mechanisms, and inference-time guidance under explicit diagnostics.
Bridging these components into reliable, scalable tools for complex geophysical simulators remains an open and high-impact direction for future work.
The field of diffusion-based SBI is rapidly evolving, and continued research is needed to refine these methods, explore their combinations, and validate their performance in complex scenarios.

\newpage
\begin{appendix}
\section{Discrete-time diffusion models forward process}\label{app:discrete_diffusion_fwd}

Let $\mathbf{x}_0\in \mathbb{R}^d$ be an initial data sample at time $t=0$ drawn from the true distribution $p_{\text{data}}(\mathbf{x}_0)$ and $\mathbf{x}_t$ for $t=1,\ldots,T$ be a sequence of latent variables in the same sample space as $\mathbf{x}_0$.
At each discrete time step $t$, the forward diffusion process adds Gaussian noise to $\mathbf{x}_{t-1}$ to produce $\mathbf{x}_t$.
The Markov chain of latent variables which defines the forward noising process is given by
\begin{equation}\label{eq:discrete_forward}
    q(\mathbf{x}_{1:T} \mid \mathbf{x}_0) 
    := \prod_{t=1}^T q(\mathbf{x}_t \mid \mathbf{x}_{t-1}),
    \quad 
    q(\mathbf{x}_t \mid \mathbf{x}_{t-1}) 
    := \mathcal{N}\!\left(\sqrt{1-\beta_t}\,\mathbf{x}_{t-1}, \, \beta_t I \right).
\end{equation}
Here, $\{\beta_t\}_{t=1}^T$ is a variance schedule with small $\beta_t > 0$.
This variance schedule controls the amount of noise added at each time step.
The variable $q$ represents the forward noising process, often referred to as the {diffusion} process.

Each step of added Gaussian noise preserves the Markov property, meaning that $\mathbf{x}_t$ depends only on $\mathbf{x}_{t-1}$.
Iterating over~\eqref{eq:discrete_forward} yields a closed-form expression for the marginal distribution of $\mathbf{x}_t$ in terms of $\mathbf{x}_0$:
\begin{equation}\label{eq:discrete_marginal}
    q(\mathbf{x}_t \mid \mathbf{x}_0) = 
    \mathcal{N}\!\left(\sqrt{\bar{\alpha}_t}\,\mathbf{x}_0, \, (1-\bar{\alpha}_t) I \right),
    \quad 
    \bar{\alpha}_t := \prod_{s=1}^t (1-\beta_s)
\end{equation}
where $\bar{\alpha}_t$ is the cumulative product of $(1-\beta_s)$ up to time $t$.
As $t$ increases, $\bar{\alpha}_t$ decreases, causing the mean of the distribution to shrink toward zero and the variance to increase.
This mathematically signifies that as more noise is added over time, the influence of the original data $\mathbf{x}_0$ diminishes.

\section{Equivalence of score matching and noise prediction}\label{app:equivalence}

Assume the forward transition kernel is Gaussian
\[
\pi(\mathbf{x}_t \mid \mathbf{x}_0) 
= \mathcal{N}\big(\mathbf{x}_t;\,\mu(t)\mathbf{x}_0,\,{\sigma(t)}^2 \mathbf{I}\big),
\]
so that
\[
\mathbf{x}_t = \mu(t)\mathbf{x}_0 + \sigma(t)\boldsymbol{\epsilon}, 
\qquad \boldsymbol{\epsilon} \sim \mathcal{N}(\mathbf{0},\mathbf{I}).
\]
The conditional score of the forward kernel is
\begin{align}\label{eq:cond-score}
\nabla_{\mathbf{x}_t}\log p(\mathbf{x}_t \mid \mathbf{x}_0) 
&= \nabla_{\mathbf{x}_t}\Big[-\tfrac{1}{2{\sigma(t)}^2}\|\mathbf{x}_t-\mu(t)\mathbf{x}_0\|_2^2\Big] \\
&= -\frac{\mathbf{x}_t-\mu(t)\mathbf{x}_0}{{\sigma(t)}^2} \\
&= -\frac{1}{\sigma(t)}\,\boldsymbol{\epsilon}.
\end{align}

\noindent
Plugging this into the DSM loss
\[
\mathcal{L}_{\text{score}}(\phi)
= \mathbb{E}_{\mathbf{x}_0,t,\mathbf{x}_t}\!\left[
{\sigma(t)}^2 \big\| \mathbf{s}_\phi(\mathbf{x}_t,t) - \nabla_{\mathbf{x}_t}\log p(\mathbf{x}_t\mid\mathbf{x}_0)\big\|_2^2
\right],
\]
we obtain
\[
\mathcal{L}_{\text{score}}(\phi)
= \mathbb{E}\!\left[
{\sigma(t)}^2 \big\| \mathbf{s}_\phi(\mathbf{x}_t,t) + \tfrac{1}{\sigma(t)}\boldsymbol{\epsilon}\big\|_2^2
\right].
\]

\noindent
Define the noise-prediction network
\[
\boldsymbol{\epsilon}_\phi(\mathbf{x}_t,t) := -\,\sigma(t)\,\mathbf{s}_\phi(\mathbf{x}_t,t).
\]
Then \(\mathbf{s}_\phi(\mathbf{x}_t,t) = -\tfrac{1}{\sigma(t)}\boldsymbol{\epsilon}_\phi(\mathbf{x}_t,t)\), and the loss simplifies to
\begin{align*}
\mathcal{L}_{\text{score}}(\phi)
&= \mathbb{E}\!\left[
{\sigma(t)}^2 \cdot \tfrac{1}{{\sigma(t)}^2} \,
\big\| \boldsymbol{\epsilon}_\phi(\mathbf{x}_t,t)-\boldsymbol{\epsilon}\big\|_2^2
\right] \\
&= \mathbb{E}_{\mathbf{x}_0,t,\boldsymbol{\epsilon}}\!\left[
\big\| \boldsymbol{\epsilon}_\phi(\mu(t)\mathbf{x}_0+\sigma(t)\boldsymbol{\epsilon},t)-\boldsymbol{\epsilon}\big\|_2^2
\right].
\end{align*}

\noindent
Thus, under the Gaussian forward process and weighting by \({\sigma(t)}^2\), the DSM objective is algebraically equivalent to the $\epsilon$-prediction mean-squared error.

\section{Description of missing data categories}\label{appendix:missing_data}%

\begin{enumerate}
    \item Missing completely at random (MCAR): The probability of data being missing does not depend on any observed or missing values in the dataset.
    Mathematically, $p(R = 1 \mid X, \phi) = P(R = 1 \mid \phi)$ for all $X,\phi$, where $R$ is the indicator matrix for missing values and $\phi$ denotes unobserved variables. ($X$ denotes the complete data that would have been observed in the absence of missingness; $\bfx$ is then a realization of $X$.)
    \item Missing at random (MAR): The probability of missing data depends on observed values in the dataset but not on the missing data itself. 
    Mathematically, $p(R = 1 \mid \bfx, \phi) = p(R = 1 \mid \bfx_\text{o}, \phi)$ for all $\bfx_\text{mis}, \phi$. 
    Both MCAR and MAR are often considered ``ignorable'' missing data mechanisms due to the randomness in their missingness.
    \item Missing not at random (MNAR), or not missing at random (NMAR): The probability of missingness depends on the missing values themselves, or on both missing and observed values. 
    Mathematically, $p(R = 1 \mid \bfx, \phi) = p(R = 1 \mid \bfx_\text{mis}, \phi)$ or $p(R = 1 \mid \bfx_\text{o}, \bfx_\text{mis}, \phi)$. MNAR is a ``non-ignorable'' missing data mechanism.
\end{enumerate}

\section{Considerations when selecting diffusion-based SBI architectures}\label{app:comp_complexity}

\begin{table}[hbt!]
\centering
\caption{Architecture selection guide for common SBI scenarios.}
\label{tab:arch_selection}
\begin{tabular*}{\textwidth}{@{\extracolsep{\fill}} p{3.5cm} p{3cm} p{4.5cm}}
\hline
\textbf{Scenario} & \textbf{Best architecture} & \textbf{Rationale} \\
\hline
Standard SBI, many $\bfx_{\mathrm{o}}$
  & Conditional
  & Fully amortized with a single training phase \\
Single target $\bfx_{\mathrm{o}}$, tight simulation budget
  & Sequential
  & Most simulation-efficient for a single query \\
Multiple i.i.d.\ observations
  & Factorized
  & Linear scaling in $n$ without retraining \\
New prior at test time
  & Guided
  & Zero additional simulations required \\
Real-time inference required
  & Consistency
  & $100\text{--}1000\times$ faster sampling \\
Irregular or missing data
  & Conditional + attention
  & Flexible conditioning on unstructured observations \\
Function-valued parameters
  & Conditional + Fourier neural operator
  & Discretization-invariant inference \\
Correlated observations
  & Conditional (not Factorized)
  & Factorization assumes conditional independence \\
\hline
\end{tabular*}
\end{table}

\begin{sidewaystable}[p]
\centering
\caption{Practical implementation checklist across SBI architectures.
\checkmark~= required; O = optional; $\times$ = not applicable.}
\label{tab:pract_imp}
\begin{tabular*}{\textwidth}{@{\extracolsep{\fill}}
  l c c c c c}
\hline
\textbf{Component}
  & \textbf{Conditional}
  & \textbf{Guided}
  & \textbf{Sequential}
  & \textbf{Factorized}
  & \textbf{Consistency} \\
\hline
Score network $s_{\phi}$
  & \checkmark & \checkmark & \checkmark & \checkmark & $\times$ \\
Prior score $s_{\mathrm{prior}}$
  & O & \checkmark & O & \checkmark & $\times$ \\
Guidance $\nabla\log g$
  & $\times$ & \checkmark & $\times$ & $\times$ & $\times$ \\
Proposal $q^{(r)}$
  & $\times$ & $\times$ & \checkmark & $\times$ & $\times$ \\
ODE solver
  & O & O & O & O & \checkmark \\
Summary network $h(\bfx)$
  & O & $\times$ & O & O & O \\
Truncation
  & $\times$ & $\times$ & O & $\times$ & $\times$ \\
\hline
\end{tabular*}
\end{sidewaystable}

\clearpage
\begin{sidewaystable}[p]
\centering
\caption{Computational complexity comparison across diffusion-based SBI
architectures. \textbf{Legend:}
$N$: number of training simulations;
$E$: training epochs;
$T$: reverse diffusion steps;
$R$: sequential rounds (2--5);
$N_r$: simulations per round;
$n$: number of observations;
$d$: parameter dimension;
$d_g$: guidance computation cost;
$|s_{\phi}|$: score network size;
$|f_{\phi}|$: consistency model size.}
\label{tab:comp_complexity}
\begin{tabular*}{\textwidth}{@{\extracolsep{\fill}}
  l l l l l l l}
\hline
\textbf{Architecture}
  & \textbf{Train sims}
  & \textbf{Train time}
  & \textbf{Infer steps}
  & \textbf{Infer time}
  & \textbf{Mem (train)}
  & \textbf{Mem (infer)} \\
\hline
Conditional
  & $10\text{K--}100\text{K}$
  & $\mathcal{O}(N\,d\,E)$
  & $100\text{--}1000$
  & $\mathcal{O}(T\,d)$
  & $\mathcal{O}(|s_{\phi}|)$
  & $\mathcal{O}(d+|s_{\phi}|)$ \\
Guided
  & $0$ (reuse)
  & $0$
  & $100\text{--}1000$
  & $\mathcal{O}(T(d+d_g))$
  & ---
  & $\mathcal{O}(d+|s_{\phi}|)$ \\
Sequential
  & $R{\times}N_r\approx10\text{K--}50\text{K}$
  & $\mathcal{O}(R\,N_r\,d\,E)$
  & $100\text{--}1000$
  & $\mathcal{O}(T\,d)$
  & $R{\times}\mathcal{O}(|s_{\phi}|)$
  & $\mathcal{O}(d+|s_{\phi}|)$ \\
Factorized
  & $10\text{K--}100\text{K}$
  & $\mathcal{O}(N\,d\,E)$
  & $100\text{--}1000$
  & $\mathcal{O}(T\,n\,d)$
  & $\mathcal{O}(|s_{\phi}|)$
  & $\mathcal{O}(d+|s_{\phi}|)$ \\
Consistency
  & $N+M$ (distill)
  & $\mathcal{O}((N{+}M)\,d\,E)$
  & $1\text{--}4$
  & $\mathcal{O}(d)$
  & $\mathcal{O}(|f_{\phi}|)$
  & $\mathcal{O}(d+|f_{\phi}|)$ \\
\hline
\end{tabular*}
\end{sidewaystable}
\clearpage

\clearpage
\begin{sidewaystable}[p]
\centering
\caption{Common failure modes of diffusion-based SBI architectures and
recommended mitigations.}
\label{tab:failure-modes}
\begin{tabular*}{\textwidth}{@{\extracolsep{\fill}}
  p{2.5cm} p{5.5cm} p{4.5cm}}
\hline
\textbf{Architecture} & \textbf{Failure mode} & \textbf{Mitigation} \\
\hline
Conditional
  & Poor conditioning when
    $\dim(\bfx_{\mathrm{o}})\gg\dim(\bftheta)$
  & Introduce summary networks or attention-based conditioning \\
Guided
  & Score unreliable when
    $\operatorname{supp}(g)\not\subseteq
     \operatorname{supp}(p_{\mathrm{learned}})$
  & Verify prior overlap; enforce support consistency \\
Sequential
  & Proposal collapse as $q^{(r)}$ becomes overly concentrated
  & Apply truncation or regularization of proposal distributions \\
Factorized
  & Biased posteriors under correlated observations
  & Replace factorization with joint or conditional modeling \\
Consistency
  & Distillation failure due to poorly trained base model
  & Ensure high-quality pre-training of $s_{\phi}^{\mathrm{diff}}$ \\
\hline
\end{tabular*}
\end{sidewaystable}
\clearpage

\end{appendix}

\begin{acks}[Acknowledgments]
The authors would like to thank the anonymous referees, an Associate Editor and the Editor for their consideration of this review paper.
\end{acks}

\newpage
\bibliographystyle{imsart-nameyear} % Style BST file (imsart-number.bst or imsart-nameyear.bst)
\bibliography{main}       % Bibliography file (usually '*.bib')

\end{document}